\documentclass[10pt,twocolumn,letterpaper]{article}
\pdfoutput=1
\usepackage{cvpr}
\usepackage{times}
\usepackage{epsfig}
\usepackage{graphicx}
\usepackage{amsmath}
\usepackage{amssymb}

\usepackage{subcaption}
\usepackage{color}
\newcommand{\comment}[1]{}
\newcommand{\parag}[1]{\vspace{-1mm}\paragraph{#1}}

\newcommand{\bc}{\mathbf{c}}
\newcommand{\f}{\mathbf{f}}

\newcommand{\s}{\mathbf{s}}

\usepackage{multirow}

\cvprfinalcopy 

\ifcvprfinal\fi
\begin{document}

\title{Context-Aware Crowd Counting}

\author{
	Weizhe Liu\textsuperscript{}
	\quad
	Mathieu Salzmann\textsuperscript{}
	\quad
	Pascal Fua\textsuperscript{}\\
	\textsuperscript{}Computer Vision Laboratory, \'{E}cole Polytechnique F\'{e}d\'{e}rale de Lausanne (EPFL)\\
	{\tt\small \{weizhe.liu, mathieu.salzmann, pascal.fua\}@epfl.ch}
}

\maketitle
\thispagestyle{empty}


\begin{abstract}

State-of-the-art methods for counting people in crowded scenes rely on deep networks to estimate crowd density. They typically use the same filters over the whole image or over large image patches. Only then do they estimate local scale to compensate for perspective distortion. This is typically achieved by training an auxiliary classifier to select, for predefined image patches, the best kernel size among a limited set of choices. As such, these methods are not end-to-end trainable and restricted in the scope of context they can leverage.

In this paper, we introduce an end-to-end trainable deep architecture that combines features obtained using multiple receptive field sizes and learns the importance of each such feature at {\it each} image location. 
In other words, our approach adaptively encodes the scale of the contextual information required to accurately predict crowd density. This yields an algorithm that outperforms state-of-the-art crowd counting methods, especially when perspective effects are strong.

\end{abstract}

\section{Introduction}

Crowd counting is important for applications such as video surveillance and traffic control.  In recent years, the emphasis has been on developing {\it counting-by-density} algorithms that rely on regressors trained to estimate the people density per unit area so that the total number can be obtained by integration, without explicit detection being required. The regressors can be based on Random Forests~\cite{Lempitsky10}, Gaussian Processes~\cite{Chan09}, or more recently 
Deep Nets~\cite{Zhang15c,Zhang16s,Onoro16,Sam17,Xiong17,Sindagi17,Shen18,Liu18b,Li18,Sam18,Shi18,Liu18c,Idrees18,Ranjan18,Cao18}, with most state-of-the-art approaches now relying on the latter.

Standard convolutions are at the heart of these deep-learning-based approaches. By using the same filters and pooling operations over the whole image, these implicitly rely on the same receptive field everywhere. However, due to perspective distortion, one should instead change the receptive field size across the image. In the past, this has been addressed by combining either density maps extracted from image patches at different resolutions~\cite{Onoro16} or feature maps obtained with convolutional filters of different sizes~\cite{Zhang16s,Cao18}. However, by indiscriminately fusing information at all scales, these methods ignore the fact that scale varies continuously across the image. While this was addressed in~\cite{Sam17,Sam18} by training classifiers to predict the size of the receptive field to use locally, the resulting methods are not end-to-end trainable; cannot account for rapid scale changes because they assign a single scale to relatively large patches; and can only exploit a small range of receptive fields for the networks to remain of a manageable size.

In this paper, we introduce a deep architecture that explicitly extracts features over multiple receptive field sizes and learns the importance of each such feature at every image location, thus accounting for potentially rapid scale changes. In other words, our approach adaptively encodes the scale of the contextual information necessary to predict crowd density. This is in contrast to crowd-counting approaches that also use contextual information to account for scaling effects as in~\cite{Shen18}, but only in the loss function as opposed to computing true multi-scale features as we do. We will show that it works better on uncalibrated images. When calibration data is available, we will also show that it can be leveraged to infer suitable local scales even better and further increase performance.

Our contribution is therefore an approach that incorporates multi-scale contextual information directly into an end-to-end trainable crowd counting pipeline, and learns to exploit the right context at each image location. As shown by our experiments, we consistently outperform the state of the art on all standard crowd counting benchmarks, such as ShanghaiTech, WorldExpo'10, UCF\_CC\_50 and UCF\_QNRF, as well as on our own Venice dataset {\footnote{\url{https://sites.google.com/view/weizheliu/home/projects/context-aware-crowd-counting}}, which features strong perspective distortion.


\section{Related Work}
\label{sec:related}

Early crowd counting methods~\cite{Wu05,Wang11a,Lin10} tended to rely on {\it counting-by-detection}, that is, explicitly detecting individual heads or bodies  and then counting them. Unfortunately, in very crowded scenes, occlusions make detection  difficult, and these approaches have been largely displaced by {\it counting-by-density-estimation} ones, which rely on training a regressor to estimate people density in various parts of the image and then integrating.  This trend began in~\cite{Chan09,Lempitsky10,Fiaschi12}, using either Gaussian Process or Random Forests regressors. Even though approaches relying on low-level features~\cite{Chen12f,Chan08,Brostow06,Rabaud06,Chan09,Idrees13} can yield good results, they have now mostly been superseded by CNN-based methods~\cite{Zhang16s,Sam17,Cao18},  a survey of which can be found  in~\cite{Sindagi17}. The same can be said about methods that count objects instead of people~\cite{Arteta14,Arteta16,Chattopadhyay16}.

The people density we want to measure is the number of people per unit area {\it on the ground}. However, the deep nets operate in the image plane and, as a result, the density estimate can be severely affected by the local scale of a pixel, that is, the ratio between image area and corresponding ground area. This problem has long been recognized. For example, the algorithms of~\cite{Zhang15c,Kang17} use geometric information to adapt the network to different scene geometries. Because this information is not always readily available, other works have focused on handling the scale implicitly within the model. In~\cite{Sindagi17}, this was done by learning to predict pre-defined density levels. These levels, however, need to be provided by a human annotator at training time. By contrast, the algorithms of~\cite{Onoro16,Shen18} use image patches extracted at multiple scales as input to a multi-stream network. They then either fuse the features for final density prediction~\cite{Onoro16} without accounting for continuous scale changes or  introduce an {\it ad hoc} term in the training loss function~\cite{Shen18} to enforce prediction consistency across scales. This, however, does not encode contextual information into the features produced by the network and therefore has limited impact. While~\cite{Zhang16s,Cao18} aim to learn multi-scale features, by using different receptive fields, they combine all of these features to predict the density.

In other words, while the previous methods account for scale, they ignore the fact that the suitable scale varies smoothly over the image and should be handled adaptively. This was addressed in~\cite{Kang18} by weighting different density maps generated from input images at various scales. However, the density map at each scale only depends on features extracted at this particular scale, and thus may already be corrupted by the lack of adaptive-scale reasoning. Here, we argue that one should rather extract \emph{features} at multiple scales and learn how to adaptively combine them. While this, in essence, was also the motivation of~\cite{Sam17,Sam18}, which train an extra classifier to assign the best receptive field for each image patch, these methods remain limited in several important ways. First, they rely on classifiers, which requires pre-training the network before training the classifier, and thus is not end-to-end trainable. Second, they typically assign a single scale to an {\it entire} image patch that can still be large and thus do not account for rapid scale changes. Last, but not least, the range of receptive field sizes they rely on remains limited in part because using much larger ones would require using much deeper architectures, which may not be easy to train given the kind of networks being used. 

By contrast, in this paper, we introduce an end-to-end trainable architecture that adaptively fuses multi-scale features, without explicitly requiring defining patches, but rather by learning how to weigh these features for each individual pixel, thus allowing us to accommodate rapid scale changes. By leveraging multi-scale pooling operations, our framework can cover an arbitrarily large range of receptive fields, thus enabling us to account for much larger context than with the multiple receptive fields used by the above-mentioned methods. In Section~\ref{sec:results}, we will demonstrate  that it delivers superior performance.

\section{Approach}
\label{sec:approach}

As discussed above, we aim to exploit context, that is, the large-scale consistencies that often appear in images. However, properly assessing what the scope and extent of this context should be in images that have undergone perspective distortion is a challenge. To meet it,  we introduce a new deep net architecture that adaptively encodes multi-level contextual information into the features it produces. We then show how to use these scale-aware features to regress to a final density map, both when the cameras are not calibrated and when they are.

\subsection{Scale-Aware Contextual Features}
\label{sec:features}

We formulate crowd counting as regressing a people density map from an image. Given a set of $N$ training images $\{I_{i}\}_{1 \leq i \leq N}$ with corresponding ground-truth density maps $\{D_{i}^{gt}\}$, our goal is to learn a non-linear mapping $\mathcal{F}$ parameterized by $\theta$ that maps an input image $I_{i}$ to an estimated density map $D_{i}^{est}(I_{i}) = \mathcal{F}(I_{i},\theta)$ that is as similar as possible to $D_{i}^{gt}$ in $L^2$ norm terms. 

Following common practice~\cite{Long15a,Ren15,Liu16a}, our starting point is a network comprising the first ten layers of a pre-trained  \textit{VGG-16} network~\cite{Simonyan15}. Given an image $I$, it outputs features of the form
\begin{equation}
\f_{v} = \mathcal{F}_{vgg}(I) \; , \label{eq:vggFeatures}
\end{equation}
which we take as base features to build our scale-aware ones. 


\begin{figure*}
\centering
        \includegraphics[width=.9\linewidth]{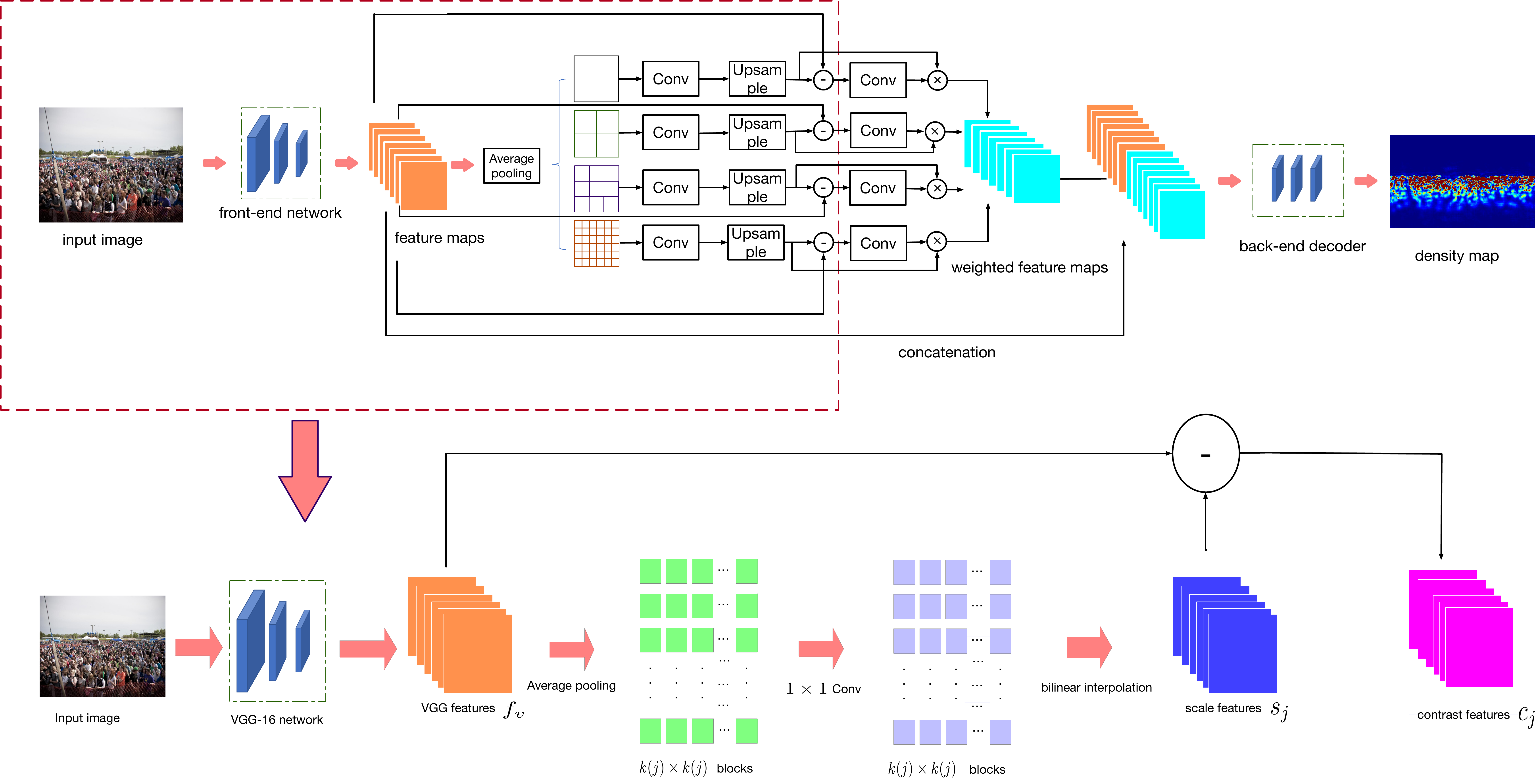}
  \vspace{-2mm}
  \caption{ {\bf Context-Aware Network.} {\bf (Top)} RGB images are fed to a font-end network that comprises the first 10 layers of the \textit{VGG-16} network. The resulting local features are grouped in blocks of different sizes by average pooling followed by a 1$\times$1 convolutional layer. They are then up-sampled back to the original feature size to form the contrast features. Contrast features are further used to learn the weights for the scale-aware features that are then fed to a back-end network to produce the final density map.  {\bf (Bottom)} As shown in this expanded version of the first part of the network, the contrast features are the difference between local features and context features.}
  \label{fig:sac}
  \end{figure*}

As discussed in Section~\ref{sec:related}, the limitation of $\mathcal{F}_{vgg}$ is that it encodes the same receptive field over the entire image. To remedy this, we compute scale-aware features by performing \textit{Spatial Pyramid Pooling}~\cite{He14b} to extract multi-scale context information from the VGG features of Eq.~\ref{eq:vggFeatures}. Specifically, as illustrated at the bottom of Fig.~\ref{fig:sac}, we compute these scale-aware features as  
\begin{equation}
\s_{j} = U_{bi}(\mathcal{F}_{j}(P_{ave}(\f_{v},j),\theta_{j})) \; , \label{eq:contextFeatures}
\end{equation}
where, for each scale $j$, $P_{ave}(\cdot,j)$ averages the VGG features into $k(j) \times k(j)$ blocks; $\mathcal{F}_{j}$ is a convolutional network with kernel size 1 to combine the context features across channels without changing their dimensions. We do this because SPP keeps each feature channel independent, thus limiting the representation power. We verified that without this the performance drops. This is in contrast to earlier arthitectures that convolve to reduce the dimension~\cite{Szegedy15,Zhao17b}; and $U_{bi}$ represents bilinear interpolation to up-sample the array of contextual features to be of the same size as $\f_{v}$. In practice, we use $S=4$ different scales, with corresponding block sizes $k(j) \in \{1,2,3,6\}$ since it shows better performance compared with other settings.

The simplest way to use our scale-aware features would be to concatenate all of them to the original VGG features $\f_{v}$. This, however, would not account for the fact that scale varies across the image. To model this, we propose to learn to predict weight maps that set the relative influence of each scale-aware feature at each spatial location. To this end, we first define contrast features as
\begin{equation}
\bc_{j} = \s_{j} - \f_{v}   \; .\label{eq:contrastFeatures} 
\end{equation}
They capture the differences between the features at a specific location and those in the neighborhood, which often is an important visual cue that denotes saliency. Note that, for human beings, saliency matters. For example, in the image of Fig.~\ref{fig:contrasted}, the eye is naturally drawn to the woman at the center in part because edges in the rest of the image all point in her direction and that edges at her location do not. In our context, these contrast features provide us with important information to understand the local scale of each image region. We therefore exploit them as input to auxiliary networks with weights $\theta_{sa}^j$ 
that compute the weights $\omega_{j}$ assigned to each one of the $S$ different scales we use. Each such network outputs a scale-specific weight map of the form
\begin{equation}
\omega_{j} = \mathcal{F}_{sa}^j(\bc_j,\theta_{sa}^j) \;. \label{eq:aux_net}
\end{equation}
$\mathcal{F}_{sa}^j$ is a 1$\times$1 convolutional layer followed by a sigmoid function to avoid division by zero. We then employ these weights to compute our final contextual features as
\begin{equation}
\f_{I} = \left[ \f_{v} | \frac{\sum_{j=1}^{S}\omega_{j} \odot \s_{j}}{\sum_{j=1}^{S}\omega_{j}}  \right] \;,  \label{eq:fi}
\end{equation}
where $\left[ \cdot | \cdot \right]$ denotes the channel-wise concatenation operation, and $\odot$ is the element-wise product between a weight map and a feature map. 

Altogether, as illustrated in Fig.~\ref{fig:sac}, the network $\mathcal{F}(I,\theta)$ extracts the contextual features $\f_{I}$ as discussed above, which are then passed to a decoder consisting of several dilated convolutions that produces the density map. The specific architecture of the network is described in Table~\ref{tab:arch}. As shown by our experiments, this network already outperforms the state of the art on all benchmark datasets, without explicitly using information about camera geometry. As discussed below, however, these results can be further improved when such information is available.


\begin{figure}
\centering
        \includegraphics[width=.9\linewidth]{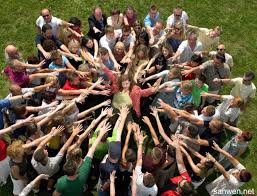}
        \vspace{-3mm}
  \caption{ {\bf Context and saliency.}  People's gaze tends be drawn to the person in the center, probably because most the image edges point in that direction.}
  \label{fig:contrasted}
  \end{figure}

\subsection{Geometry-Guided Context Learning}
\label{sec:geometry}

Because of perspective distortion, the contextual scope suitable for each region varies across the image plane.
Hence, scene geometry is highly related to contextual information and could be used to guide the network to better adjust to 
the scene context it needs.

We therefore extend the previous approach to exploiting geometry information when it is available.
To this end, we represent the scene geometry of image $I_{i}$ with a \textit{perspective map} $M_{i}$, which encodes the number of pixels per meter in the image plane. Note that this perspective map has the same spatial resolution as the input image. We therefore use it as input to a truncated \textit{VGG-16} network. In other words, the base features of Eq.~\ref{eq:vggFeatures} are then replaced by features of the form
\begin{equation}
    \f_{g} = \mathcal{F'}_{vgg}( M_{i},\theta_{g} )\;,
\end{equation}
where $\mathcal{F'}_{vgg}$ is a modified \textit{VGG-16} network with a single input channel. To initialize the weights corresponding to this channel, we average those of the original three RGB channels. Note that we also normalize the perspective map $M_{i}$ to lie within the same range as the RGB images. Even though this initialization does not bring any obvious difference in the final counting accuracy, it makes the network converge much faster.

To further propagate the geometry information to later stages of our network, we exploit the modified VGG features described above, which inherently contain geometry information, as an additional input to the auxiliary network of Eq.~\ref{eq:aux_net}. Specifically, the weight map for each scale is then computed as
\begin{equation}
    \omega_{j} = \mathcal{F}_{gc}^j(\left[\bc_{j} | \f_{g} \right],\theta_{gc}^j)\;.
\end{equation}
These weight maps are then used as in Eq.~\ref{eq:fi}. Fig.~\ref{fig:gc} depicts the corresponding architecture.


\begin{table}[t]
  \centering
  \begin{tabular}{ |c|c|c|c|}
    \hline
    layer & front-end($\mathcal{F}_{vgg}$) & layer & back-end decoder \\
    \hline
     1 - 2 & 3$\times$3$\times$64 conv-1 & 1 & 3$\times$3$\times$512 conv-2\\
    \hline
   & 2 $\times$ 2 max pooling & 2 & 3$\times$3$\times$512 conv-2\\
    \hline
     3 - 4 & 3$\times$3$\times$128 conv-1 & 3 & 3$\times$3$\times$512 conv-2\\
    \hline
    &2 $\times$ 2 max pooling  & 4  & 3$\times$3$\times$256 conv-2\\
    \hline
    5 - 7 & 3$\times$3$\times$256 conv-1 & 5 & 3$\times$3$\times$128 conv-2\\
    \hline
    &2 $\times$ 2 max pooling & 6 & 3$\times$3$\times$64 conv-2\\
    \hline
     8 - 10 & 3$\times$3$\times$512 conv-1 & 7 & 1$\times$1$\times$1 conv-1 \\
    \hline
    \end{tabular}
  \caption{{\bf Network architecture of proposed model}
  { Convolutional layers are represented as ``(kernel size) $\times$ (kernel size) $\times$ (number of filters) conv-(dilation rate)".  }}
  \label{tab:arch}
\end{table}

\subsection{Training Details and Loss Function}

Whether with or without geometry information, our networks are trained using the $L^2$ loss defined as
\begin{equation}
    L(\theta)  =  \frac{1}{2B}\sum_{i=1}^{B}\|D^{gt}_{i}-D^{est}_{i}\|^{2}_{2} \;,
    \label{eq:loss}
\end{equation}
where $B$ is the batch size. To obtain the ground-truth density maps $D^{gt}_{i}$, we rely on the same strategy as previous work~\cite{Li18,Sam17,Zhang16s,Sam18}. Specifically, 
to each image $I_{i}$, we associate a set of $c_{i}$ 2D points $P_{i}^{gt} = {\{P_{i}^j} \}_{1 \leq j \leq c_i}$ that denote the position of each human head in the scene. The corresponding ground-truth density map $D_{i}^{gt}$ is obtained by convolving an image containing ones at these locations and zeroes elsewhere with a Gaussian kernel $\mathcal{N}^{gt}(p|\mu,\sigma^{2})$~\cite{Liu18}. We write 
\begin{eqnarray}
    \forall p \in I_{i}, D_{i}^{gt}(p|I_{i})  = \sum_{j=1}^{c_i}\mathcal{N}^{gt}(p|\mu=P_{i}^j,\sigma^{2})\;, 
    \label{eq:densityMap}
\end{eqnarray}
where $\mu$ and $\sigma$ represent the mean and standard deviation of the normal distribution. To produce the comparative results we will show in Section~\ref{sec:results}, we use the same $\sigma$ as the methods we compare against.

To minimize the loss of Eq.~\ref{eq:loss}, we use Stochastic Gradient
Descent (SGD) with batch size 1 for various size dataset and Adam with batch size 32 for fixed size dataset. Furthermore, during training, we randomly crop image patches of $\frac{1}{4}$ the size of the original image at different locations. These patches are further mirrored to double the training set.


\begin{figure*}
\centering
  \includegraphics[width=.9\linewidth]{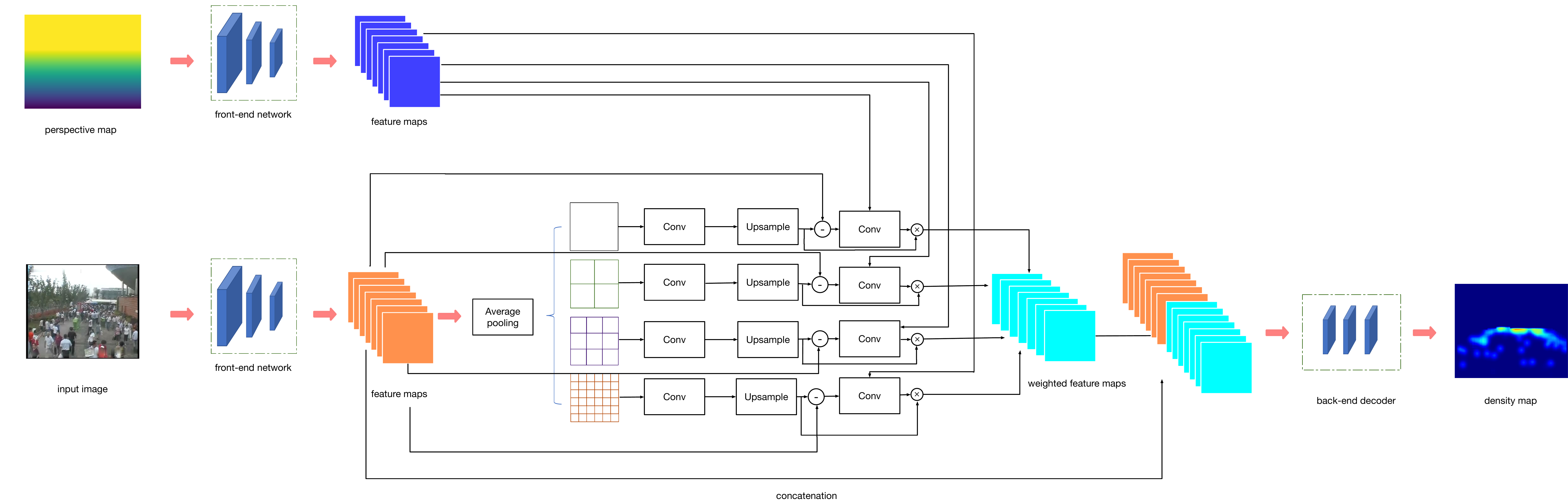}
  \vspace{-3mm}
  \caption{{\bf Expanded Context-Aware Network.} To account for camera registration information when available, we add a branch to the architecture of Fig.~\ref{fig:sac}. It takes as input a perspective map that encodes local scale. Its output is concatenated to the original contrast features and the resulting scale-aware features are used to estimate people density.}
  \label{fig:gc}
  \end{figure*}


\newcommand{\oursS}[0]{{\bf OURS-CAN}}
\newcommand{\oursG}[0]{{\bf OURS-ECAN}}

\newcommand{\vggS}[0]{{\bf VGG-SIMPLE}}
\newcommand{\vggC}[0]{{\bf VGG-CONCAT}}
\newcommand{\vggN}[0]{{\bf VGG-NCONT}}

\section{Experiments}
\label{sec:results}

In this section, we evaluate the proposed approach. We first introduce the evaluation metrics and benchmark datasets we use in our experiments. We then compare our approach to state-of-the-art methods, and finally perform a  detailed ablation study.

\subsection{Evaluation Metrics}
\label{sec:metrics}

Previous works in crowd density estimation use the mean absolute error ($MAE$) and the root mean squared error ($RMSE$) as evaluation metrics~\cite{Zhang16s,Zhang15c,Onoro16,Sam17,Xiong17,Sindagi17}. They are defined as  
\begin{equation}
    MAE = \frac{1}{N}\sum_{i=1}^{N}|z_{i}-\hat{z_{i}}| \mbox{ and } RMSE=\sqrt{\frac{1}{N}\sum_{i=1}^{N}(z_{i}-\hat{z_{i}})^{2}} \; , \nonumber
\end{equation}
where $N$ is the number of test images, $z_{i}$ denotes the true number of people inside the ROI of the $i$th image and $\hat{z_{i}}$ the estimated number of people. In the benchmark datasets discussed below, the ROI is the whole image except when explicitly stated otherwise. Note that number of people can be recovered by integrating over the pixels of the predicted density maps as $\hat{z_{i}} = \sum_{p \in I_{i} } D_{i}^{est}(p|I_{i})$.

\subsection{Benchmark Datasets and Ground-truth Data}

We use five different datasets to compare our approach to recent ones. The first four were released along with recent papers and have already been used for comparison purposes since. We created the fifth one ourselves and will make it publicly available as well.

\parag{ShanghaiTech~\cite{Zhang16s}.} 
It comprises 1,198 annotated images with 330,165 people in them. It is divided in part A with 482 images and part B with 716. In part A, 300 images form the training set and, in part B, 400. The remainder are used for testing purposes. For a fair comparison with earlier work~\cite{Zhang16s,Shen18,Li18,Shi18}, we created the ground-truth density maps in the same manner as they did. Specifically, for Part A, we used the geometry-adaptive kernels introduced in~\cite{Zhang16s}, and for part B, fixed kernels. In Fig.~\ref{fig:shanghai}, we show one image from each part, along with the ground-truth density maps and those estimated by our algorithm.


\begin{figure}
\centering
      \begin{subfigure}{.15\textwidth}
        \includegraphics[width=.9\linewidth]{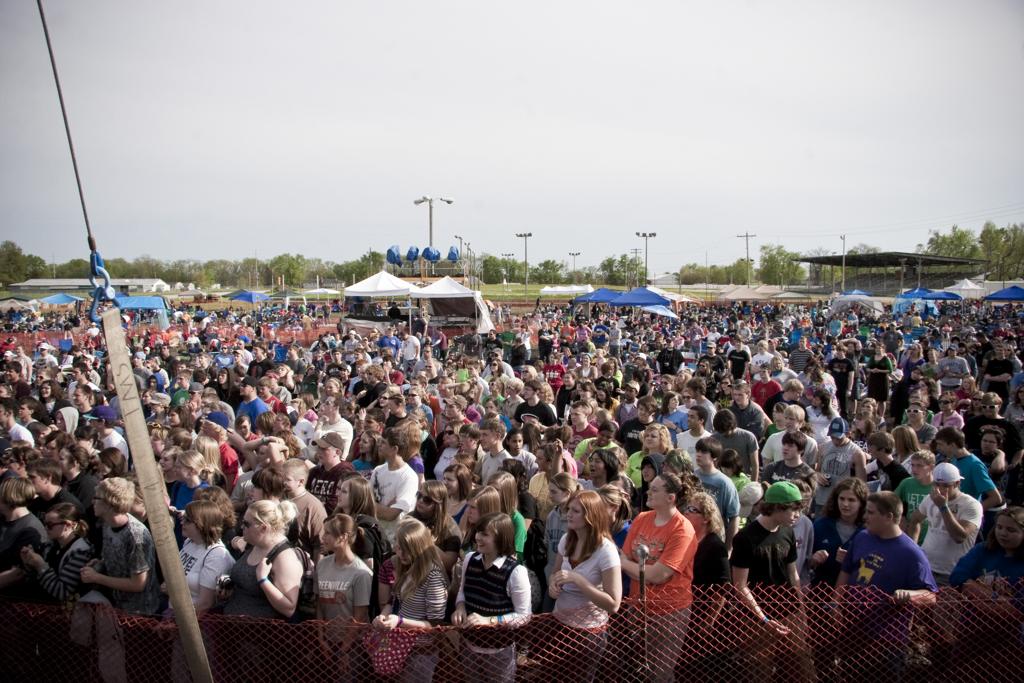}
        \includegraphics[width=.9\linewidth]{partB.jpg}
        \caption{Input image}
      \end{subfigure}\hfill
      \begin{subfigure}{.15\textwidth}
          \includegraphics[width=.9\linewidth]{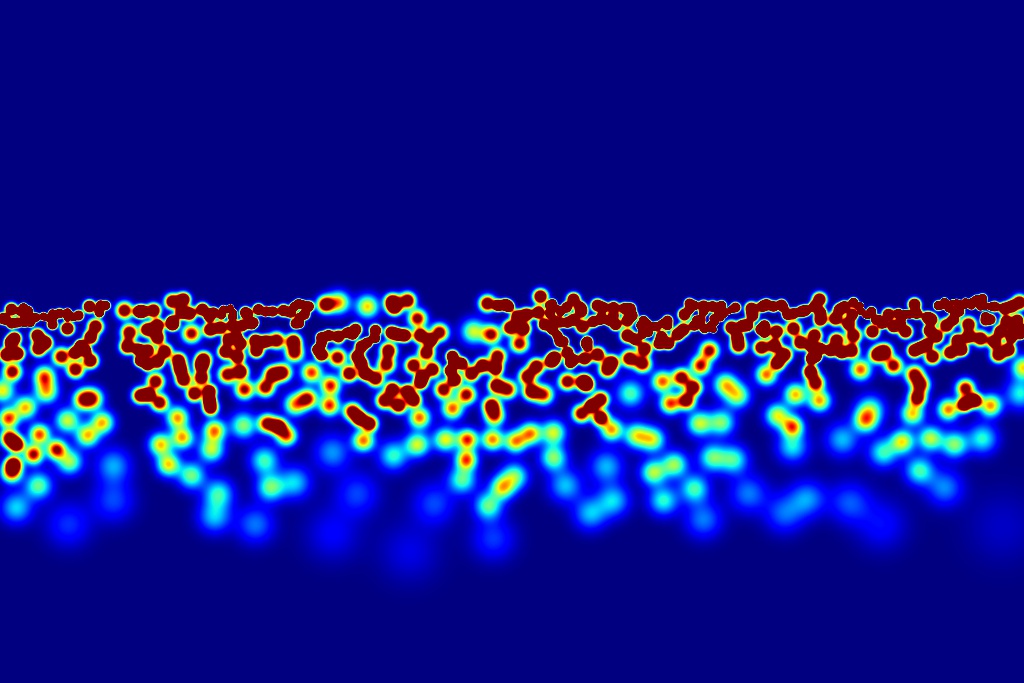}
          \includegraphics[width=.9\linewidth]{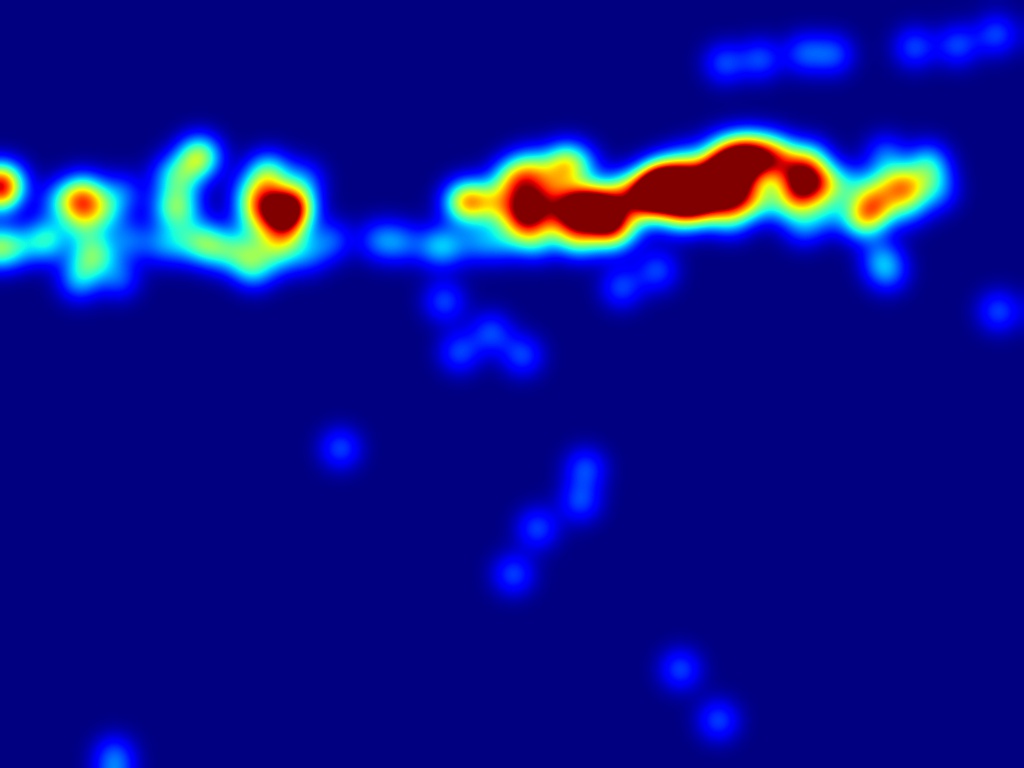}
          \caption{Ground truth}
      \end{subfigure}\hfill
      \begin{subfigure}{.15\textwidth}
          \includegraphics[width=.9\linewidth]{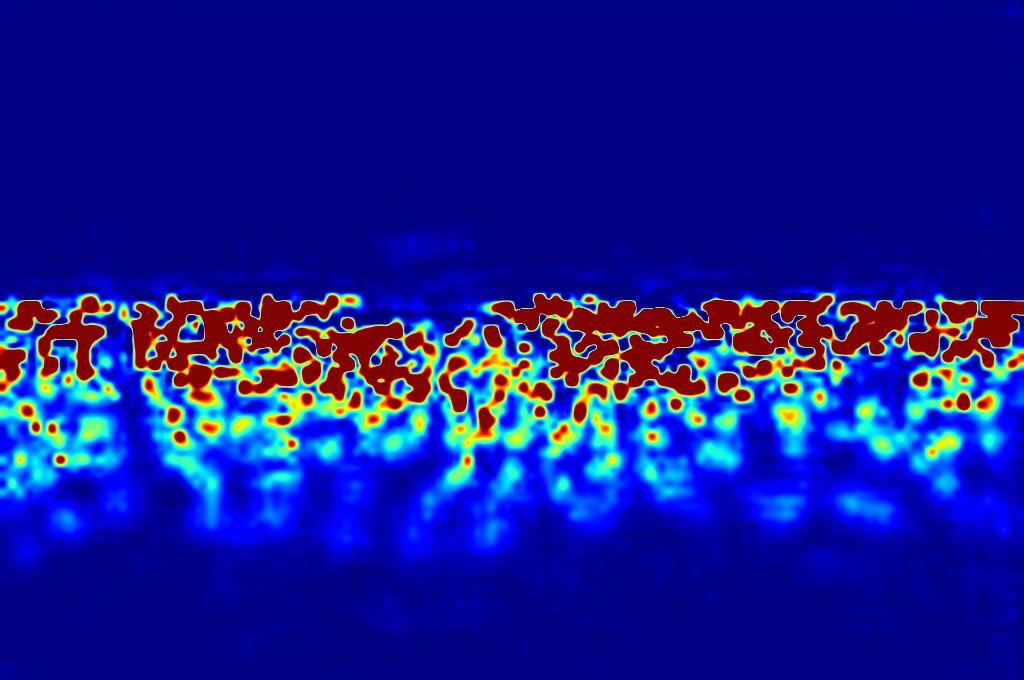}
          \includegraphics[width=.9\linewidth]{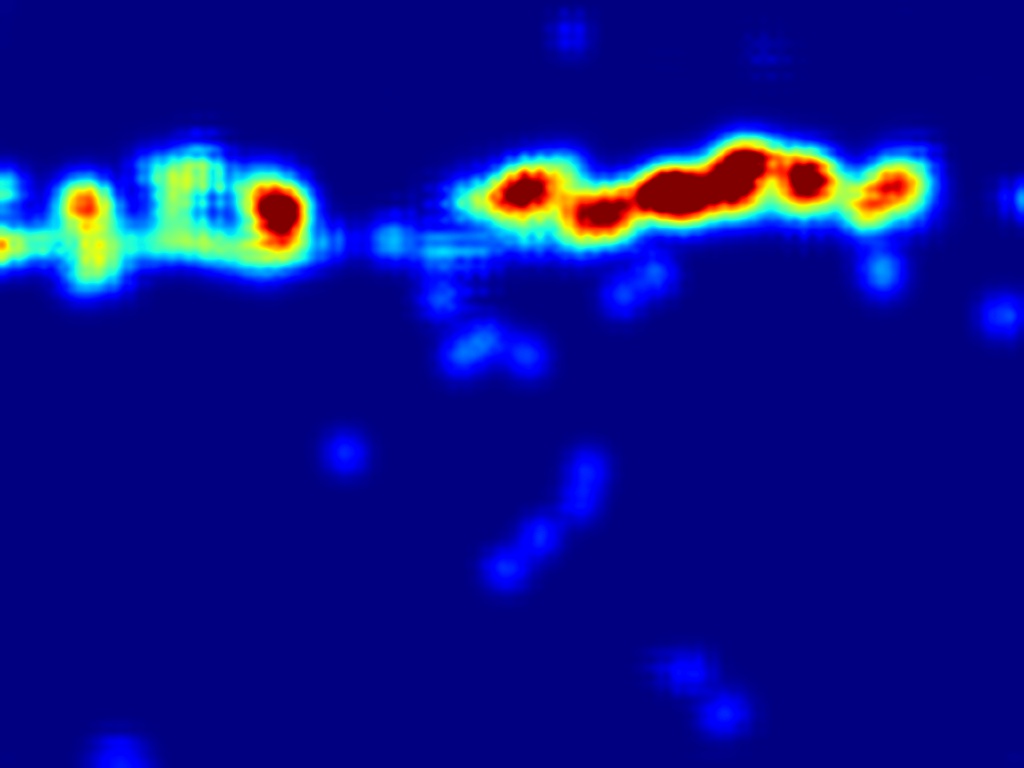}
          \caption{Our prediction}
      \end{subfigure}
  \vspace{-3mm}
  \caption{{\bf Crowd density estimation on  ShanghaiTech}. First row: Image from Part A. Second row: Image from Part B.  Our model adjusts to rapid scale changes and delivers density maps that are close to the ground truth. }
  \label{fig:shanghai}
  \end{figure}

\parag{UCF-QNRF~\cite{Idrees18}.} 
It comprises 1,535 jpeg images with 1,251,642 people in them. The training set is made of 1,201 of these images. Unlike in {\bf ShanghaiTech}, there are dramatic variations both in crowd density and image resolution. The ground-truth density maps were generated by adaptive Gaussian kernels as in~\cite{Idrees18}.

\parag{UCF\_CC\_50~\cite{Idrees13}.} 
It contains only 50 images with a people count varying from 94 to 4,543, which makes it challenging  for a deep-learning approach. For a fair comparison again, the ground-truth density maps were generated using fixed kernels and we follow the same 5-fold cross-validation  protocol as in~\cite{Idrees13}: We partition the images into 5 10-image groups. In turn, we then pick four groups for training and the remaining one for testing. This gives us 5 sets of results and we report their average.

\parag{WorldExpo'10~\cite{Zhang15c}.} 
It comprises 1,132 annotated video sequences collected from 103 different scenes. There are 3,980 annotated frames, with 3,380 of them used for training purposes.
Each scene contains a Region Of Interest (ROI)  in which people are counted. The bottom row of Fig.~\ref{fig:VeniceAndShangai} depicts three of these images and the associated camera calibration data. We generate the ground-truth density maps as in our baselines~\cite{Sam17,Li18,Cao18}. As in previous work~\cite{Zhang15c,Zhang16s,Sam17,Sam18,Li18,Cao18,Liu18,Sindagi17,Shen18,Ranjan18,Shi18} on this dataset, we report the \textit{MAE} of each scene, as well as the average over all scenes.

\parag{Venice.}
The four datasets discussed above have the advantage of being publicly available but do not contain precise calibration information. In practice, however, it can be readily obtained using either standard photogrammetry techniques or onboard sensors, for example when using a drone to acquire the images. To test this kind of scenario, we used a cellphone to film additional sequences of the Piazza San Marco in Venice, as seen from various viewpoints on the second floor of the basilica, as shown in the top two rows of Fig.~\ref{fig:VeniceAndShangai}. We then used the white lines on the ground to compute camera models. As shown in the bottom two rows of Fig.~\ref{fig:VeniceAndShangai}, this yields a more accurate calibration than in \textbf{WorldExpo'10}. The resulting dataset contains 4 different sequences and in total 167 annotated frames with fixed 1,280 $\times$ 720 resolution. 80 images from a single long sequence are taken as training data, and we use the images from the remaining 3 sequences for testing purposes. The ground-truth density maps were generated using  fixed Gaussian kernels as in part B of the {\bf ShanghaiTech} dataset. 


\begin{figure*}[htbp]
\centering
\begin{tabular}{cccc}
 \includegraphics[width=.3\linewidth]{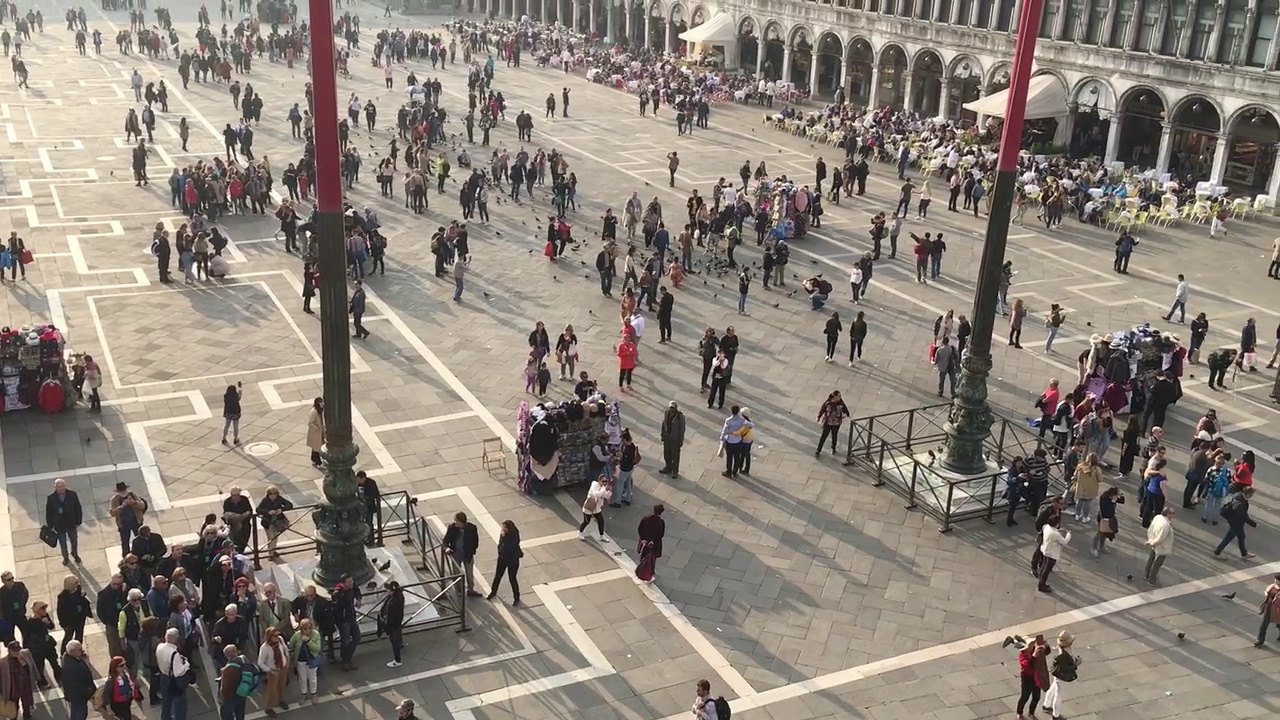}&
 \includegraphics[width=.3\linewidth]{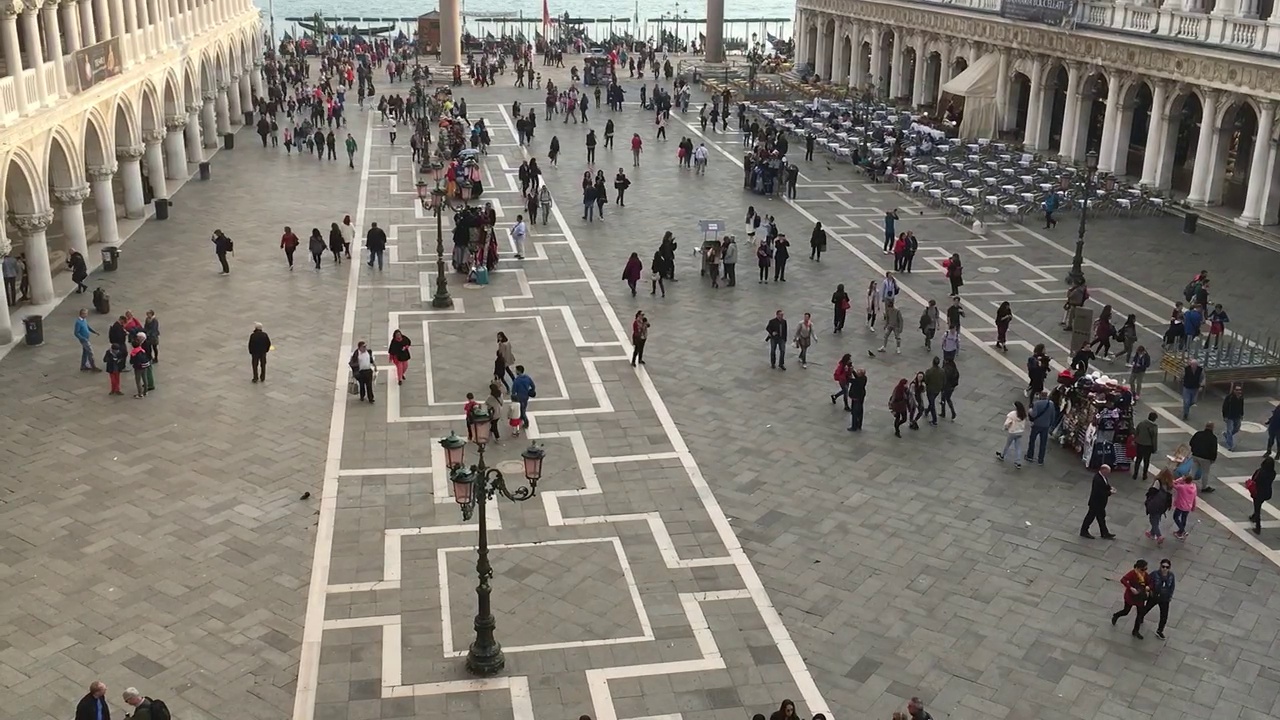}&
 \includegraphics[width=.3\linewidth]{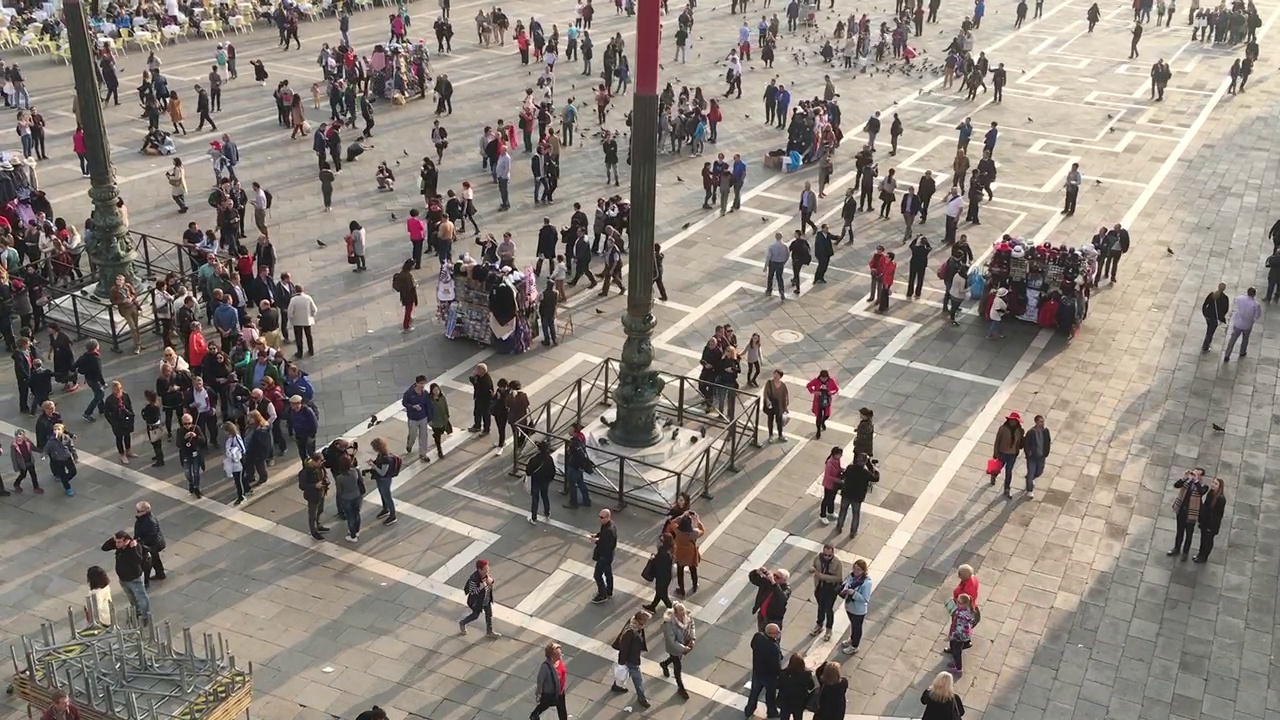}\\
 \includegraphics[width=.3\linewidth]{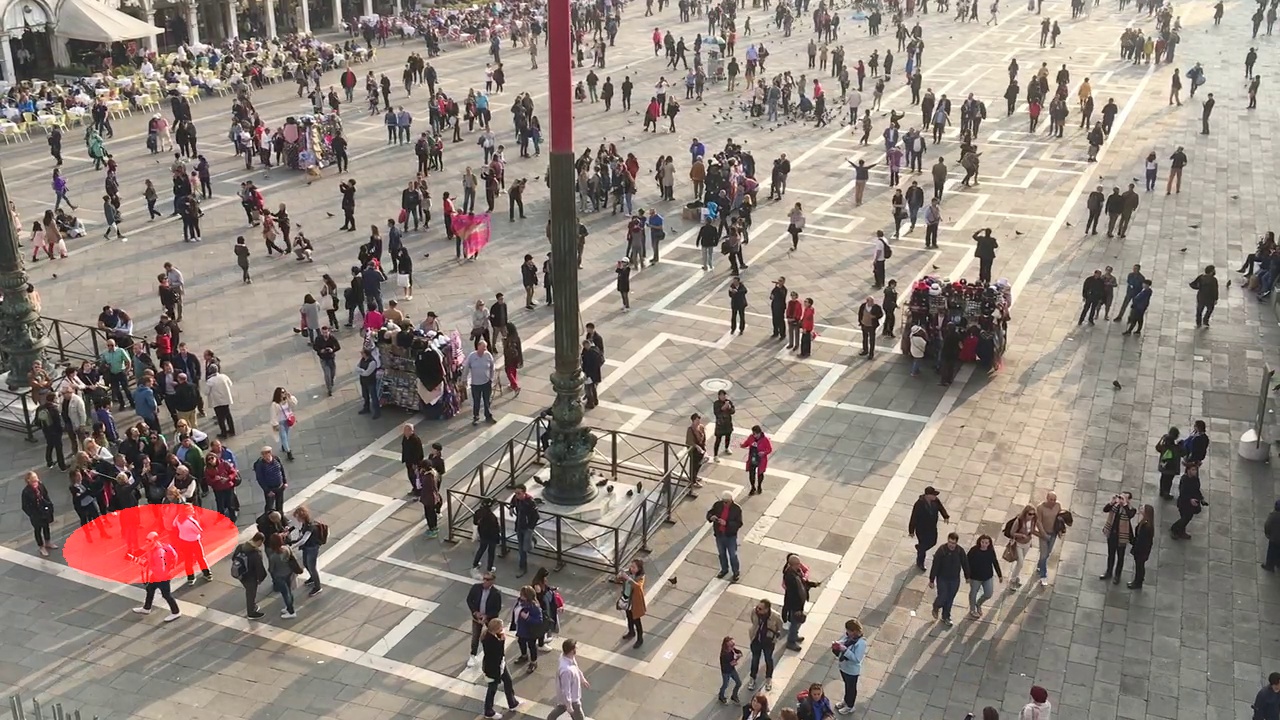}&
 \includegraphics[width=.3\linewidth]{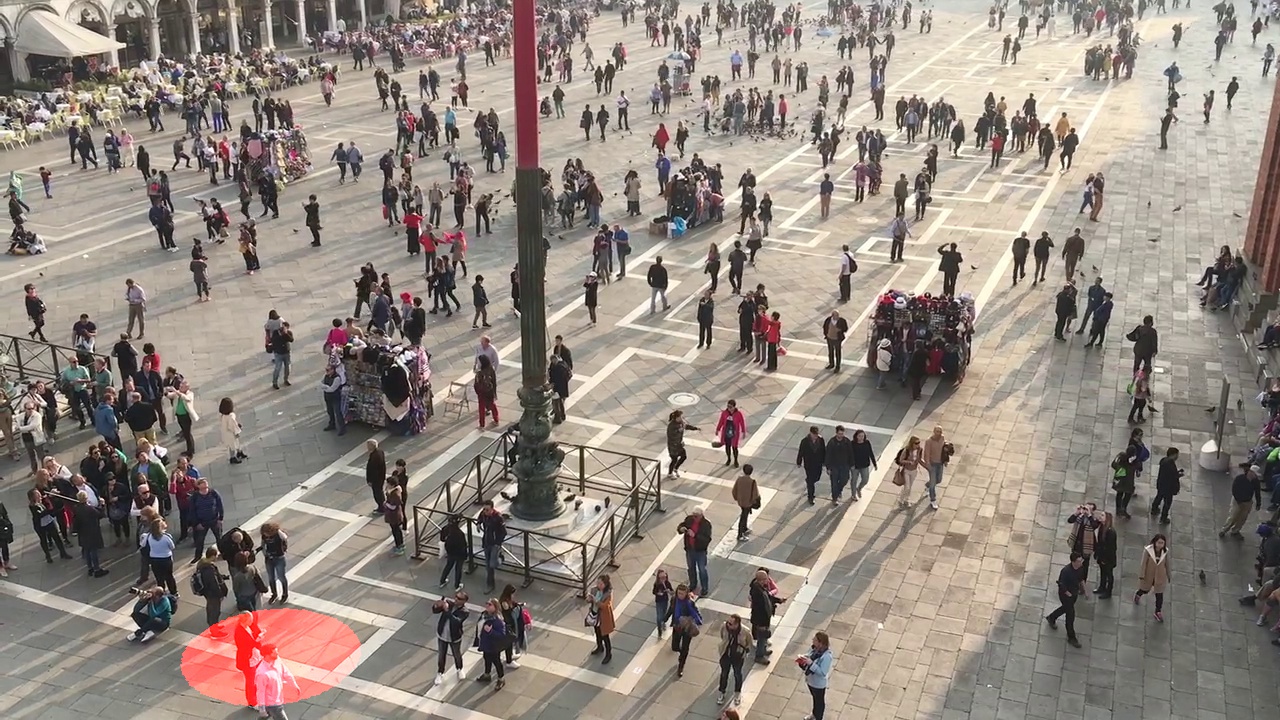}&
 \includegraphics[width=.3\linewidth]{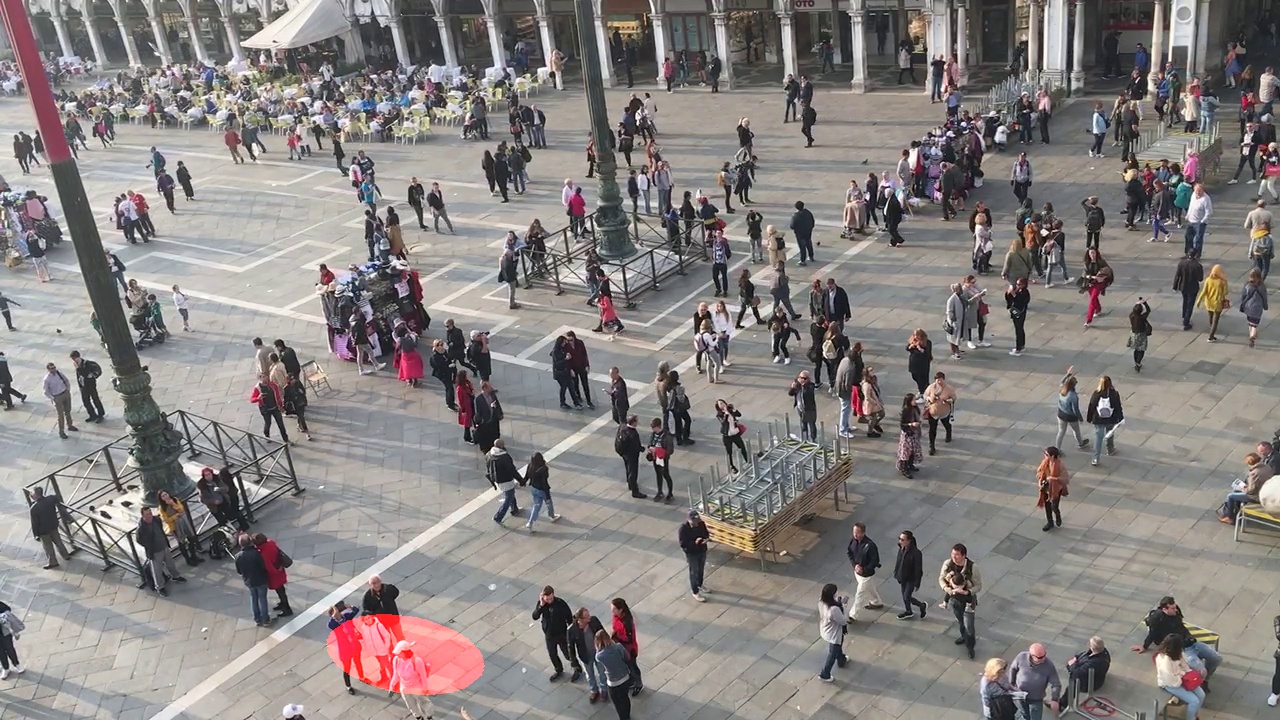}\\
 \includegraphics[width=.3\linewidth]{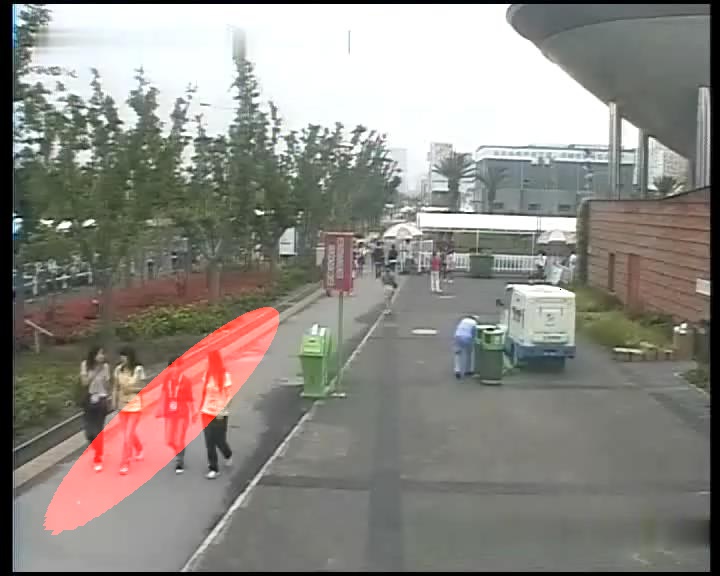}&
 \includegraphics[width=.3\linewidth]{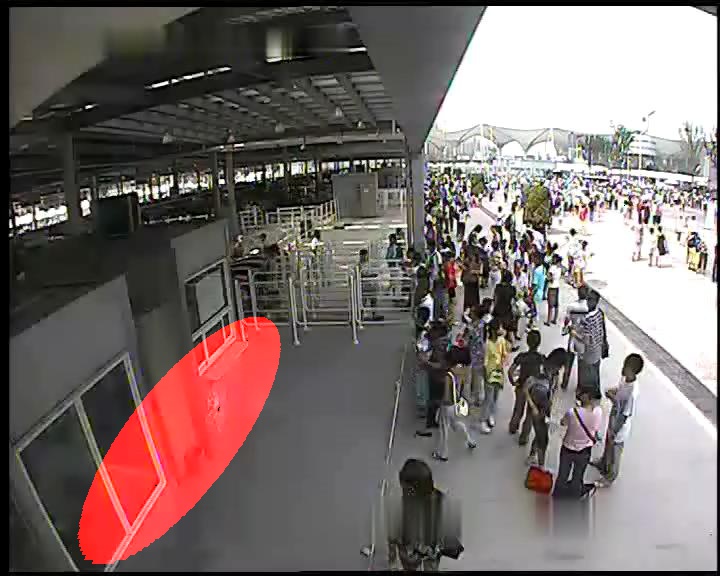}&
 \includegraphics[width=.3\linewidth]{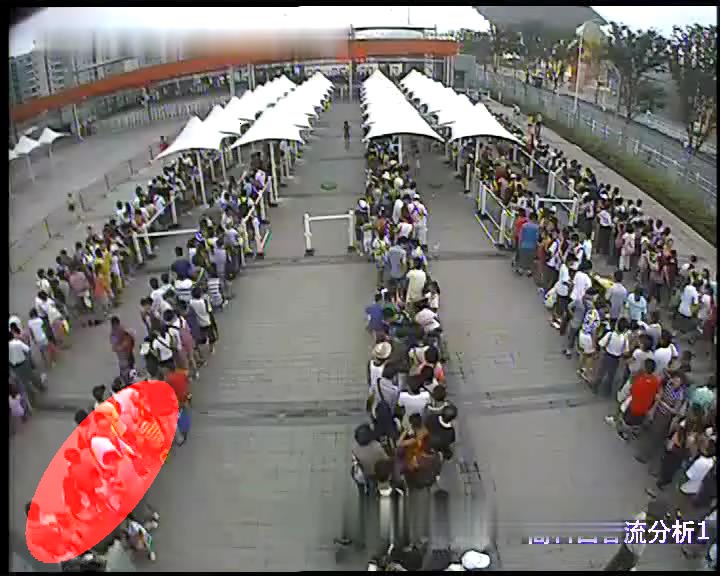}
\end{tabular}
\vspace{-3mm}
  \caption{ {\bf Calibration in Venice and WorldExpo'10.} (Top row) Images of Piazza San Marco taken from different viewpoints. (Middle row) We used the regular ground patterns to accurately register the cameras in each frame. The red ellipse overlaid in red is the projection of a 1m radius circle from the ground plane to the image plane.  
 (Bottom row) The same 1m radius circle overlaid on three WorldExpo'10 images. As can be seen in the bottom right image, the ellipse surface corresponds to an area that could be filled by many more people that could realistically fit in a 1m radius circle. By contrast, the ellipse deformations are more consistent and accurate for Venice, which denotes a better registration.
 }
  \label{fig:VeniceAndShangai}
  \end{figure*}

\subsection{Comparing against Recent Techniques}


\begin{table}
  \centering
\begin{tabular}{ |p{2.5cm}|c|c|c|c|c|c|}
  \hline
  & \multicolumn{2}{|c}{Part\_A} & \multicolumn{2}{|c|}{Part\_B} \\
  \hline
  Model  & $MAE$ & $RMSE$  & $MAE$ & $RMSE$ \\
  \hline
  Zhang \textit{et al.}~\cite{Zhang15c} & 181.8 & 277.7 & 32.0 & 49.8 \\
  \hline
  MCNN~\cite{Zhang16s} & 110.2 & 173.2 & 26.4 &41.3  \\
  \hline
  Switch-CNN~\cite{Sam17} & 90.4 & 135.0 & 21.6 & 33.4  \\
  \hline
  CP-CNN~\cite{Sindagi17}  & 73.6 & 106.4 & 20.1 & 30.1 \\
  \hline 
  ACSCP~\cite{Shen18} & 75.7 & 102.7 & 17.2 & 27.4\\
  \hline
  Liu \textit{et al.}~\cite{Liu18b} & 73.6 & 112.0 & 13.7 & 21.4\\
  \hline
  D-ConvNet~\cite{Shi18} & 73.5 & 112.3 & 18.7 & 26.0 \\ 
  \hline
  IG-CNN~\cite{Sam18} & 72.5 & 118.2 & 13.6 & 21.1\\
  \hline 
  ic-CNN\cite{Ranjan18} & 68.5 & 116.2 & 10.7 & 16.0\\
  \hline
  CSRNet~\cite{Li18} & 68.2 & 115.0 & 10.6 & 16.0\\
  \hline
  SANet~\cite{Cao18} & 67.0 & 104.5 & 8.4 & 13.6\\
  \hline
  \oursS{} & \textbf{62.3} & \textbf{100.0} & \textbf{7.8} & \textbf{12.2} \\
  \hline
  \end{tabular}
   \vspace{-3mm}
  \caption{ {\bf Comparative results on the ShanghaiTech dataset.} }
  \label{tab:shanghai}
\end{table}

\begin{table}
  \centering
\begin{tabular}{ |p{3cm}|c|c|c|c|c|c|}
  \hline
  Model  & $MAE$ & $RMSE$   \\
  \hline 
  Idrees \textit{et al.}~\cite{Idrees13} & 315 & 508 \\
  \hline
  MCNN~\cite{Zhang16s} & 277 & 426 \\
  \hline
  Encoder-Decoder~\cite{Badrinarayanan15} & 270 & 478 \\
  \hline
  CMTL~\cite{Sindagi17b} & 252 & 514 \\
  \hline
  Switch-CNN~\cite{Sam17} & 228 & 445  \\
  \hline
  Resnet101~\cite{He16} & 190 & 277 \\
  \hline 
  Densenet201~\cite{Huang17c} & 163 & 226 \\
  \hline 
  Idrees \textit{et al.}~\cite{Idrees18} & 132 & 191 \\
  \hline
  \oursS{} & \textbf{107} & \textbf{183}  \\
  \hline
  \end{tabular}
  \vspace{-3mm}
  \caption{ {\bf Comparative results on the UCF\_QNRF dataset.} }
  \label{tab:ucfqnrf}
\end{table}

\begin{table}
  \centering
\begin{tabular}{ |p{3cm}|c|c|c|c|c|c|}
  \hline
  Model  & $MAE$ & $RMSE$   \\
  \hline 
  Idrees \textit{et al.}\cite{Idrees13} & 419.5 & 541.6 \\
  \hline
  Zhang \textit{et al.}~\cite{Zhang15c} & 467.0 & 498.5  \\
  \hline
  MCNN~\cite{Zhang16s} & 377.6 & 509.1 \\
  \hline
  Switch-CNN~\cite{Sam17} & 318.1 & 439.2  \\
  \hline
  CP-CNN~\cite{Sindagi17}  & 295.8 & 320.9  \\
  \hline 
  ACSCP~\cite{Shen18} & 291.0 & 404.6 \\
  \hline
  Liu \textit{et al.}~\cite{Liu18b} & 337.6 & 434.3 \\
  \hline
  D-ConvNet~\cite{Shi18} & 288.4 & 404.7 \\ 
  \hline
  IG-CNN~\cite{Sam18} & 291.4 & 349.4 \\
  \hline 
  ic-CNN\cite{Ranjan18} & 260.9 & 365.5\\
  \hline
  CSRNet~\cite{Li18} & 266.1 & 397.5 \\
  \hline
  SANet~\cite{Cao18} & 258.4 & 334.9 \\
  \hline
  \oursS{} & \textbf{212.2} & \textbf{243.7}  \\
  \hline
  \end{tabular}
  \vspace{-3mm}
  \caption{ {\bf Comparative results on the UCF\_CC\_50 dataset.} }
  \label{tab:ucf50}
\end{table}

\begin{table}
  \centering
\tabcolsep=0.02cm
\begin{tabular}{ |p{2.5cm}|c|c|c|c|c|c|}
  \hline
  Model  & Scene1 & Scene2 & Scene3 & Scene4 & Scene5 &Average \\
  \hline
  Zhang \textit{et al.}~\cite{Zhang15c} & 9.8 & 14.1 & 14.3 & 22.2 & 3.7 & 12.9  \\
  \hline
  MCNN~\cite{Zhang16s} & 3.4 & 20.6 & 12.9 & 13.0 & 8.1 & 11.6 \\
  \hline
  Switch-CNN~\cite{Sam17} & 4.4 & 15.7 & 10.0 & 11.0 & 5.9 & 9.4  \\
  \hline
  CP-CNN~\cite{Sindagi17}  & 2.9 & 14.7 & 10.5 & 10.4 & 5.8 & 8.9  \\
  \hline 
  ACSCP~\cite{Shen18} & 2.8 & 14.05 & 9.6 & 8.1 & 2.9 & 7.5 \\
  \hline
  IG-CNN~\cite{Sam18} & 2.6 & 16.1 & 10.15 & 20.2 & 7.6 & 11.3 \\
  \hline 
  ic-CNN\cite{Ranjan18} & 17.0 & 12.3 & 9.2 & 8.1 & 4.7 & 10.3\\
  \hline
  D-ConvNet~\cite{Shi18} & \textbf{1.9} & 12.1 & 20.7 & 8.3 & \textbf{2.6} & 9.1\\ 
  \hline
  CSRNet~\cite{Li18} & 2.9 & 11.5 & \textbf{8.6} & 16.6 & 3.4 & 8.6 \\
  \hline
  SANet~\cite{Cao18} & 2.6 & 13.2 & 9.0 & 13.3 & 3.0 & 8.2 \\
  \hline
  DecideNet~\cite{Liu18} & 2.0 & 13.14 & 8.9 & 17.4 & 4.75 & 9.23 \\
  \hline
  \oursS{} & 2.9 & 12.0 & 10.0 & \textbf{7.9} & 4.3 & \textbf{7.4}  \\
  \hline
  \oursG{} & 2.4 & \textbf{9.4} & 8.8 & 11.2  & 4.0 & \textbf{7.2}  \\
  \hline
  \end{tabular}
  \vspace{-3mm}
  \caption{ {\bf Comparative results in MAE terms on the WorldExpo'10 dataset. }}
  \label{tab:worldexpo}
\end{table}

In Tables~\ref{tab:shanghai}, \ref{tab:ucfqnrf}, \ref{tab:ucf50}, and~\ref{tab:worldexpo}, we compare our results to those of the method that returns the best results for each one of the 4 public datasets, as currently reported in the literature. They are those of~\cite{Cao18},~\cite{Idrees18},~\cite{Cao18}, and~\cite{Shen18}, respectively. In each case, we reprint the results as given in these papers and add those of \oursS{}, that is, our method as described in Section~\ref{sec:features}. On the first three datasets, we consistently and clearly outperform all other methods. On the \textbf{WorldExpo'10} dataset, we also outperform them on average, but not in every scene. More specifically, in Scenes 2 and 4 that are crowded, we do very well. By contrast, the crowds are far less dense in Scenes 1 and 5. This makes context less informative and our approach still performs honorably but looses its edge compared to the others. Interestingly, as can be seen in Table~\ref{tab:worldexpo}, in such uncrowded scenes, a detection-based method such as DecideNet~\cite{Liu18} becomes competitive whereas it isn't in the more crowded ones. In Fig.~\ref{fig:veniceCompare}, we use a  \textbf{Venice} image to show how well our approach does compared to the others in the crowded parts of the scene. 


\begin{figure}[htbp]
\centering
\begin{tabular}{cccc}
 \hspace{-4mm}\includegraphics[width=.245\linewidth]{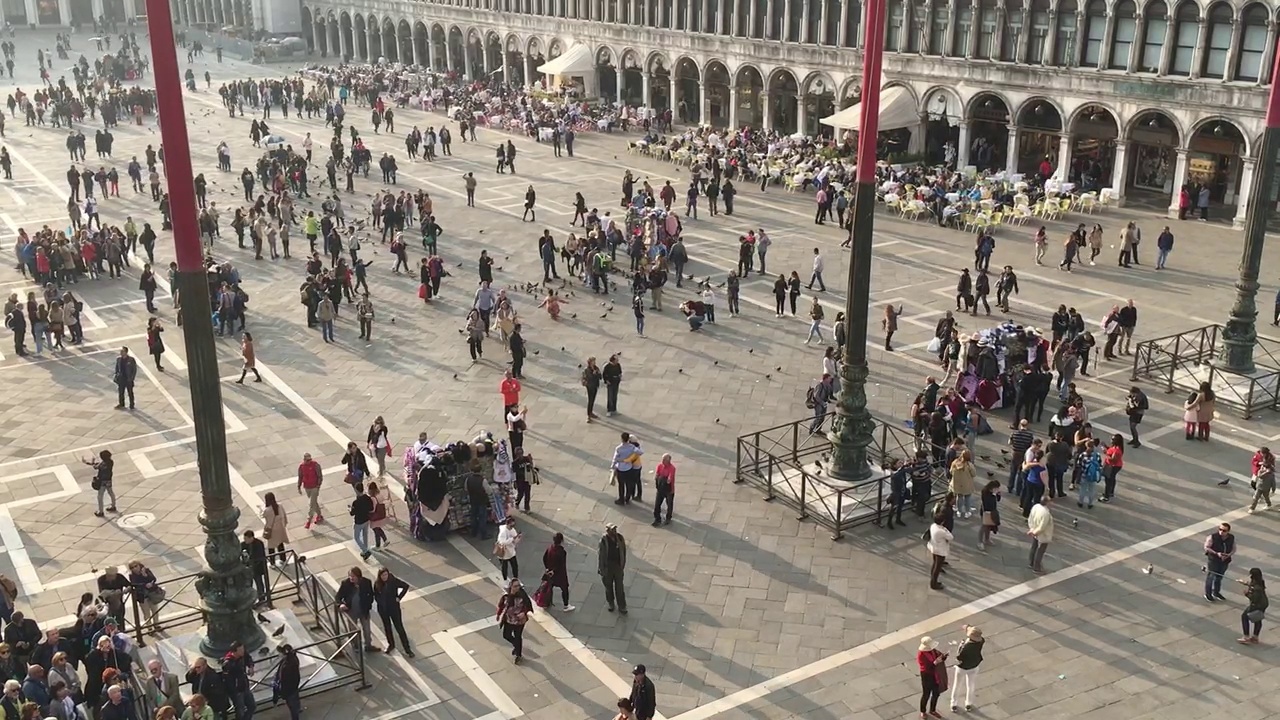}&
 \hspace{-4mm} \includegraphics[width=.245\linewidth]{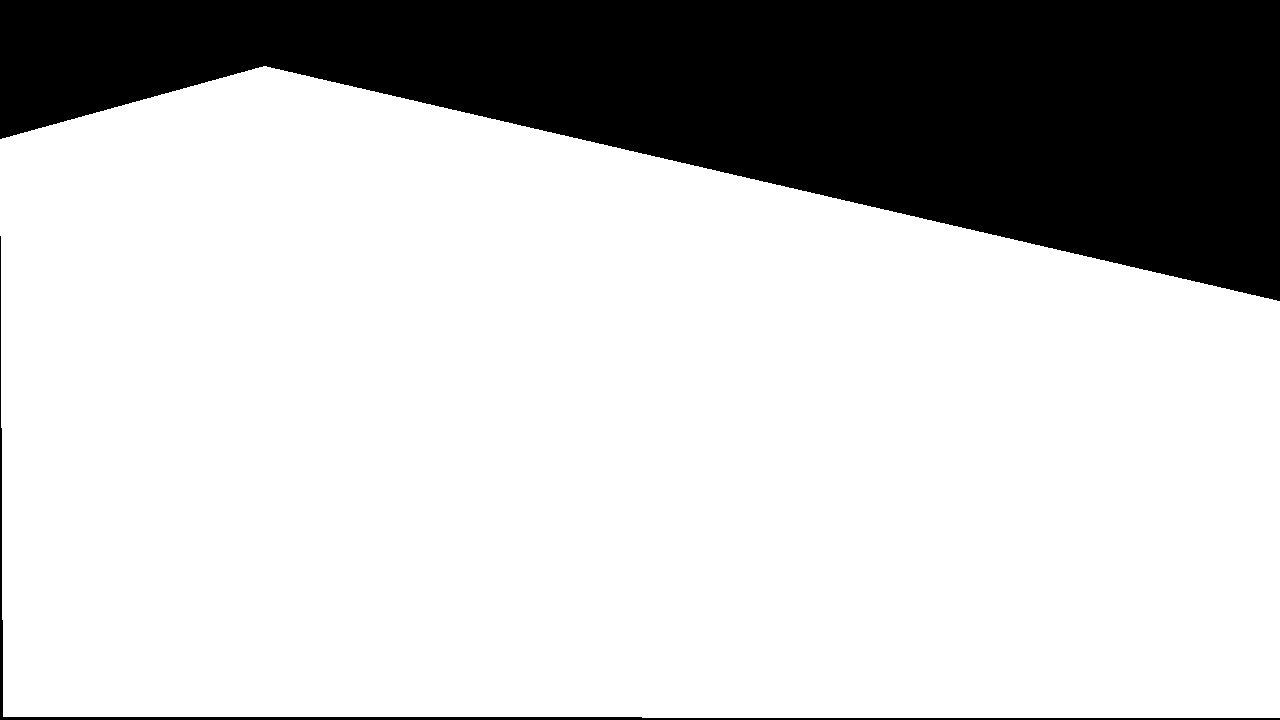}&
 \hspace{-4mm} \includegraphics[width=.245\linewidth]{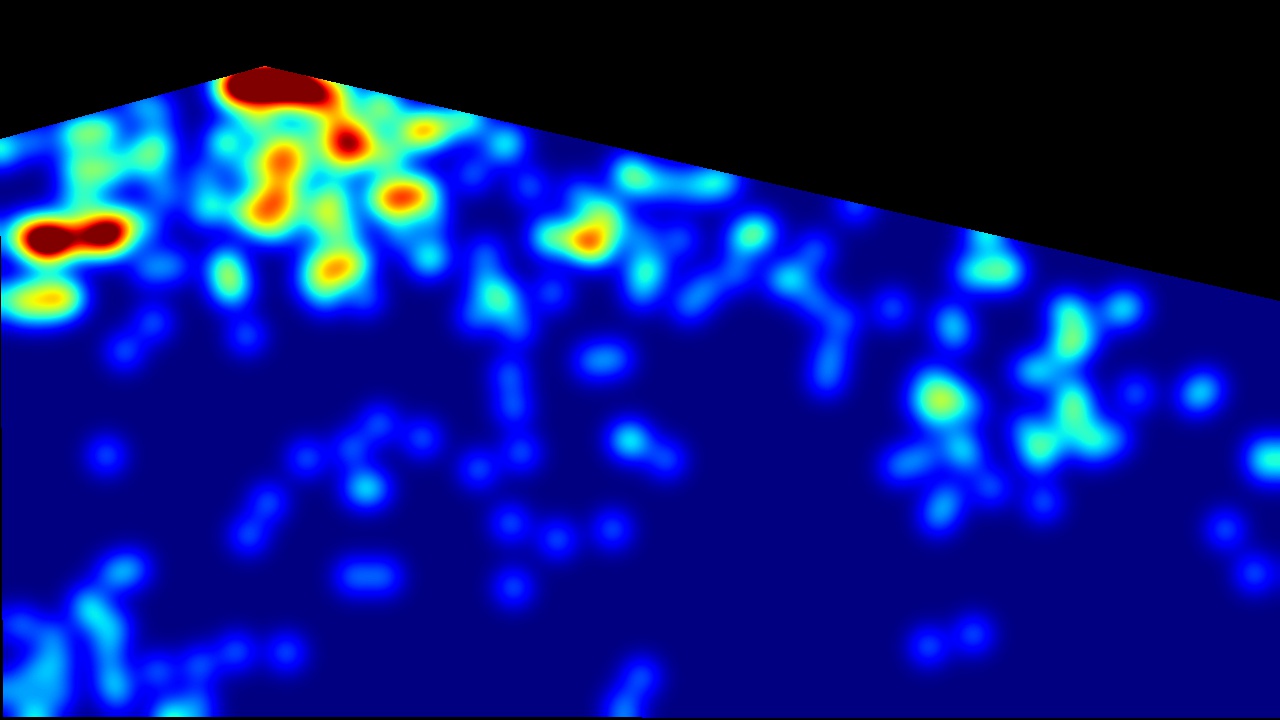}&
 \hspace{-4mm}\includegraphics[width=.245\linewidth]{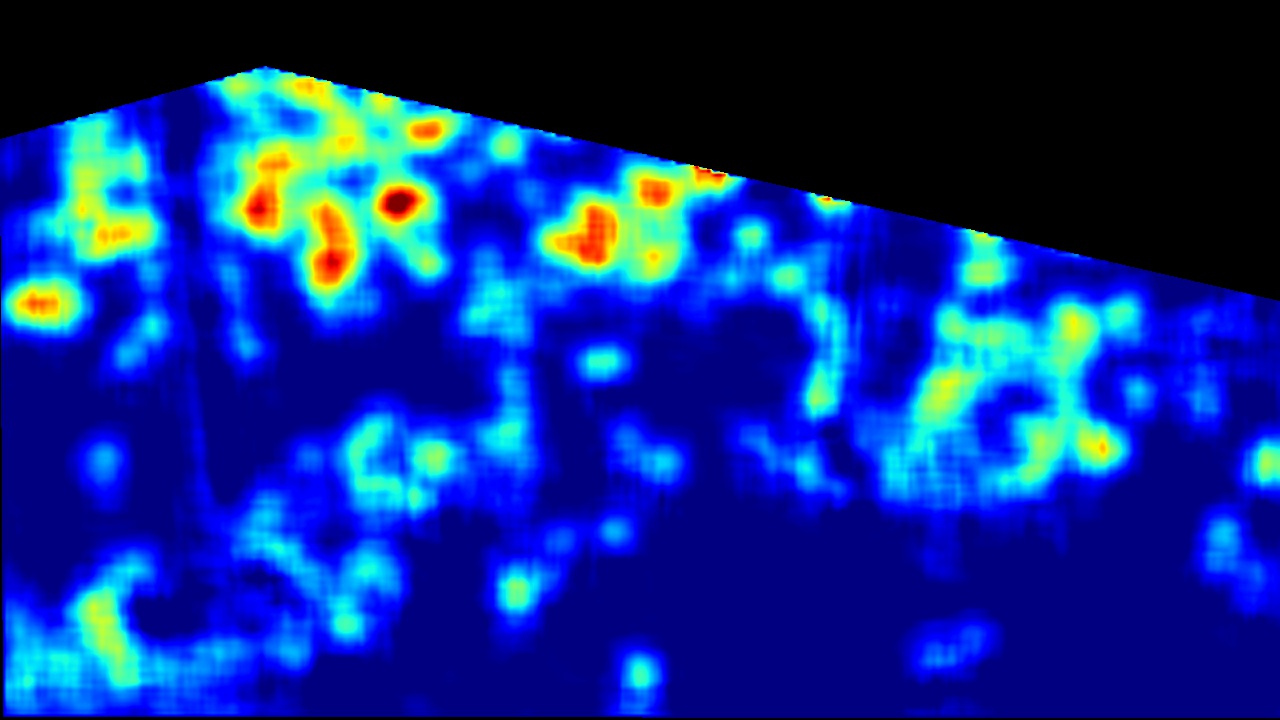}\\[-1mm]
 \hspace{-4mm}\footnotesize{Original image}&
 \hspace{-4mm}\footnotesize{Region of interest} &
 \hspace{-4mm}\footnotesize{Ground truth}&
 \hspace{-4mm}\footnotesize{\textit{MCNN}~\cite{Zhang16s}}\\[1mm]
 \hspace{-4mm}\includegraphics[width=.245\linewidth]{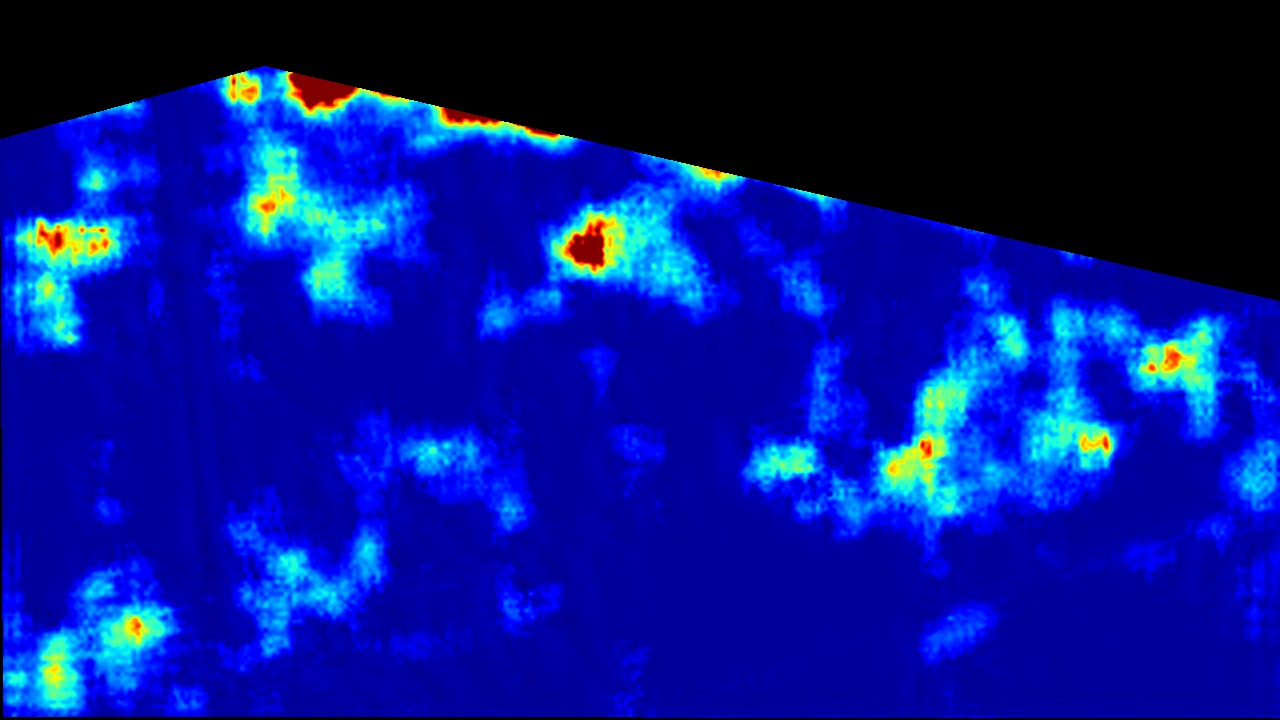}&
 \hspace{-4mm}\includegraphics[width=.245\linewidth]{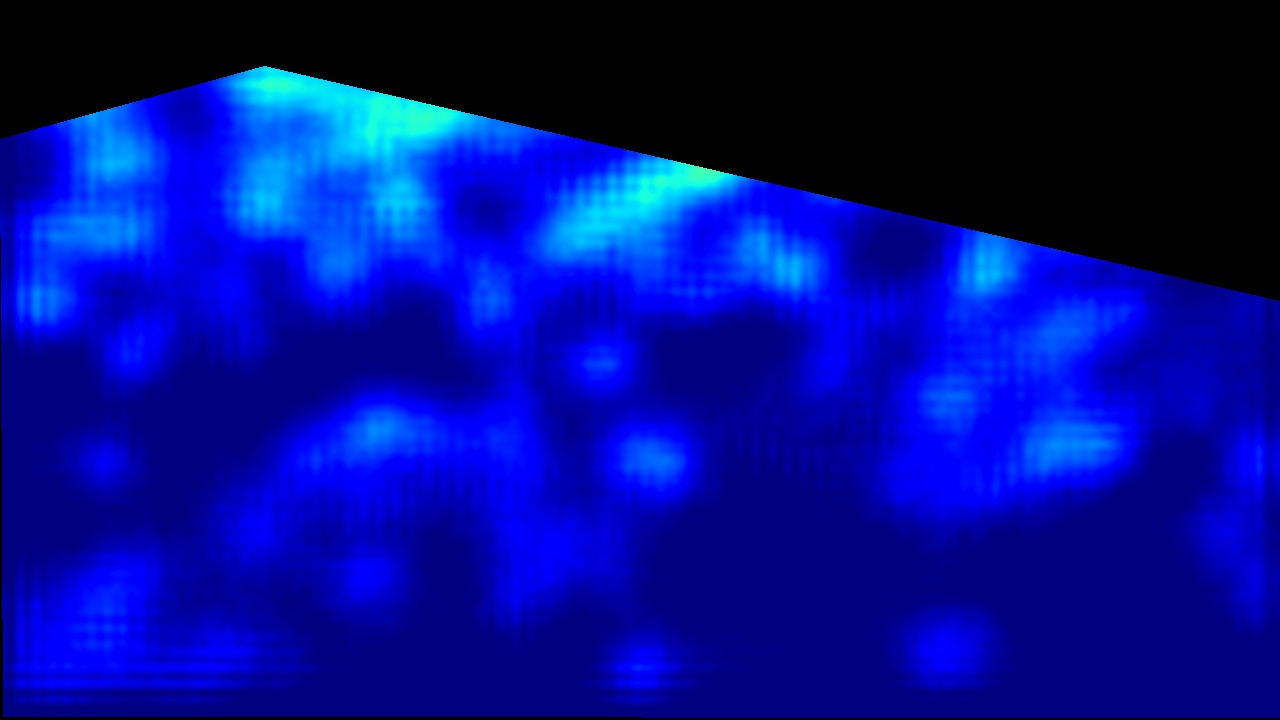}&
 \hspace{-4mm} \includegraphics[width=.245\linewidth]{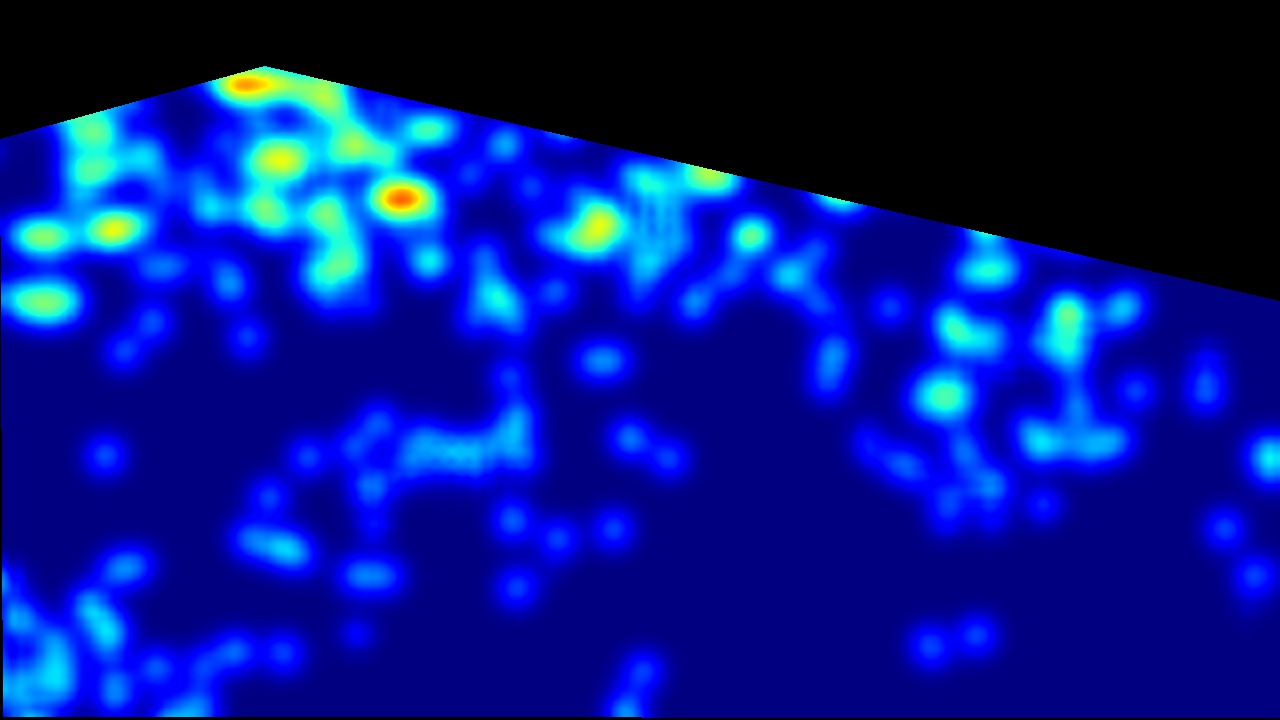}&
 \hspace{-4mm} \includegraphics[width=.245\linewidth]{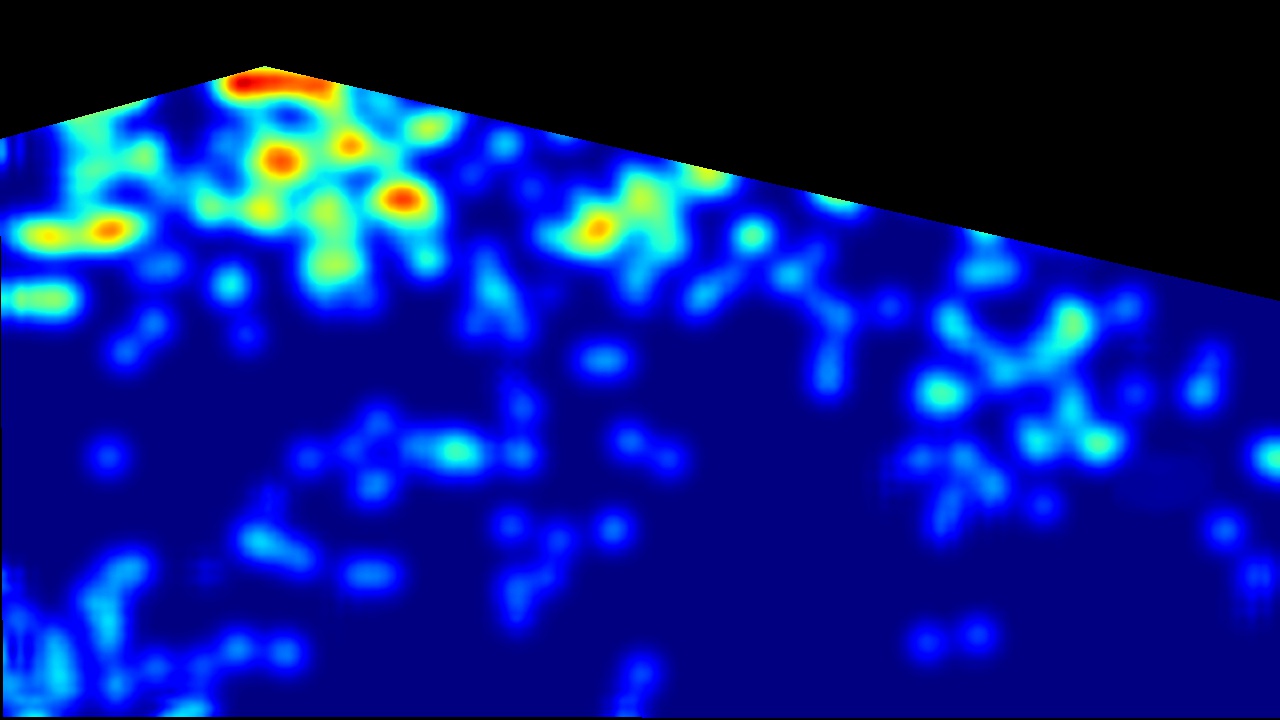}\\[-1mm]
 \hspace{-4mm}\footnotesize{\textit{Switch-CNN}~\cite{Sam17}}&
 \hspace{-4mm}\footnotesize{\textit{CSRNet}~\cite{Li18}} &
 \hspace{-4mm}\footnotesize{\oursS{}} & 
 \hspace{-4mm}\footnotesize{\oursG{}}
\end{tabular}
\vspace{-3mm}
  \caption{ {\bf Density estimation in Venice.}  Original image, ROI, ground truth density map within the ROI, and density maps estimated both by the baselines and our method. Note how much more similar the density map produced by \oursG{} is to the ground truth than the others, especially in the upper corner of the ROI, where people density is high.}
  \label{fig:veniceCompare}
  \end{figure}

The first three datasets do not have any associated camera calibration data, whereas~{\bf WorldExpo'10} comes with a rough estimation of the image plane to ground plane homography and {\bf Venice} with an accurate one. We therefore used these homographies to run \oursG{}, our method as described in  Section~\ref{sec:geometry}. We report the results in Tables~\ref{tab:worldexpo} and~\ref{tab:venice}.  Unsurprisingly, \oursG{} clearly further improves on \oursS{} when the calibration data is accurate as for {\bf Venice} and even when it is less so as for~{\bf WorldExpo}, but by a smaller margin.

\begin{figure*}[htbp]
\centering
\begin{tabular}{ccccc}
	\includegraphics[width=.18\linewidth]{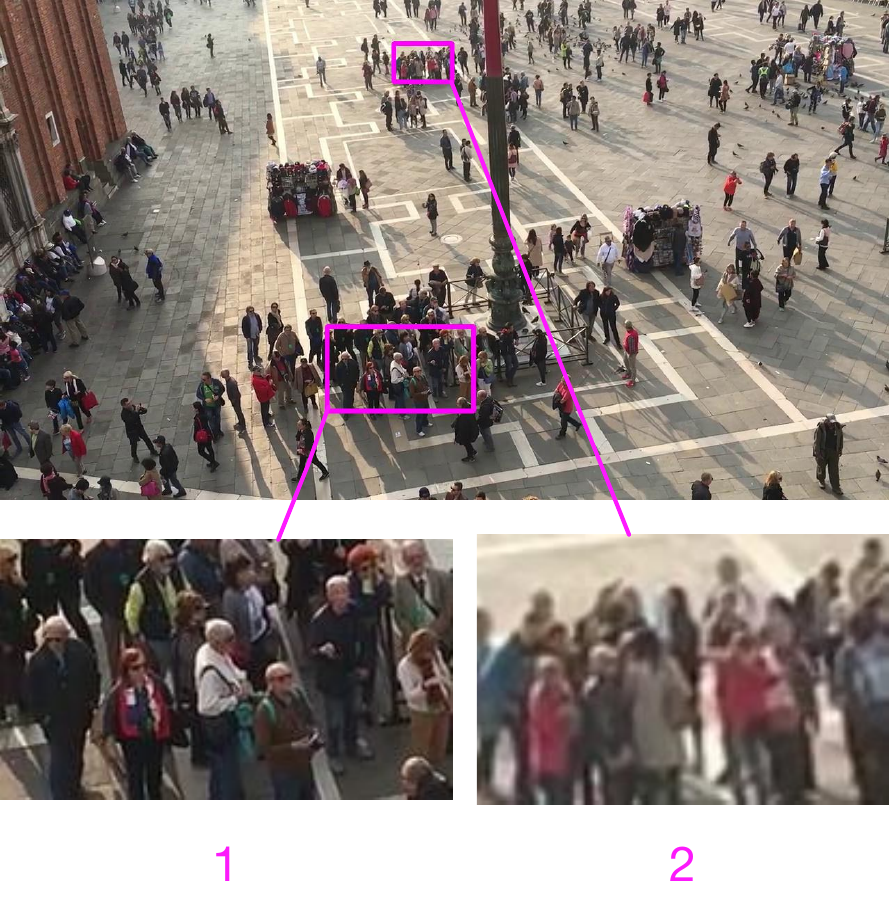}&
	\includegraphics[width=.18\linewidth]{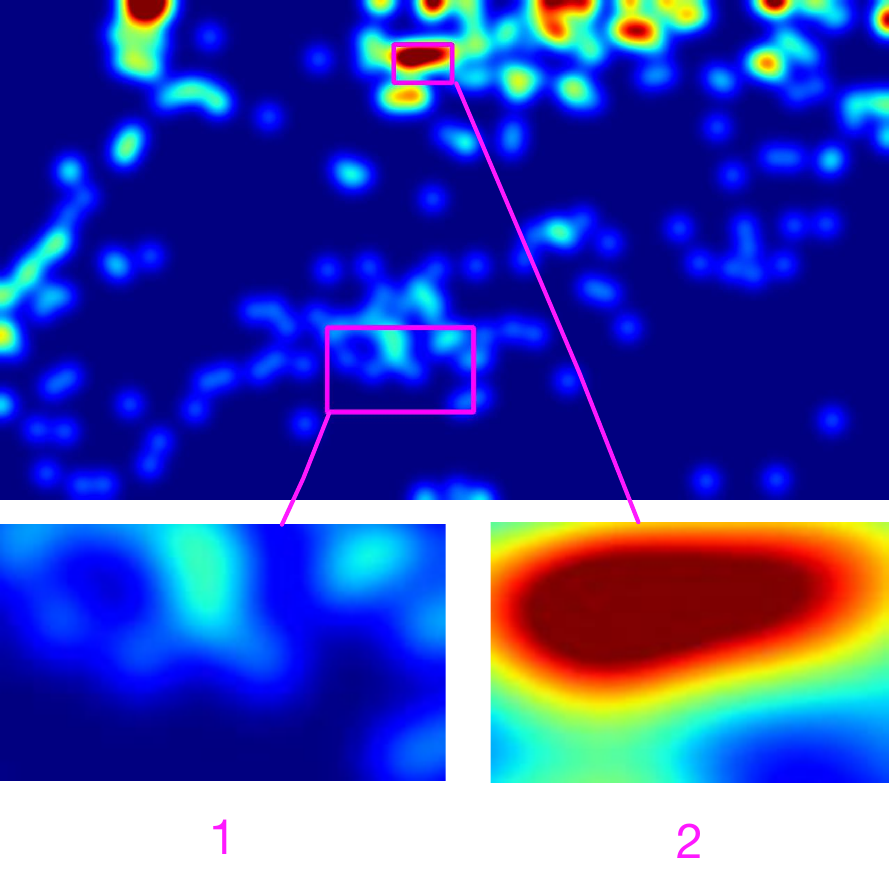}&
	\includegraphics[width=.18\linewidth]{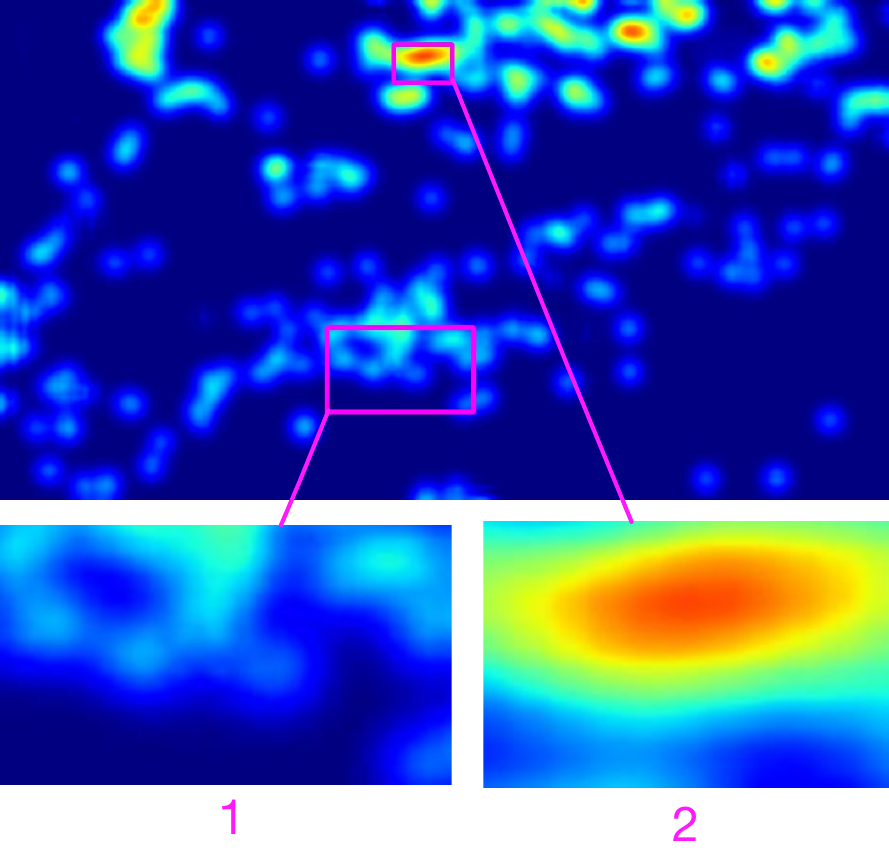}&
	\includegraphics[width=.18\linewidth]{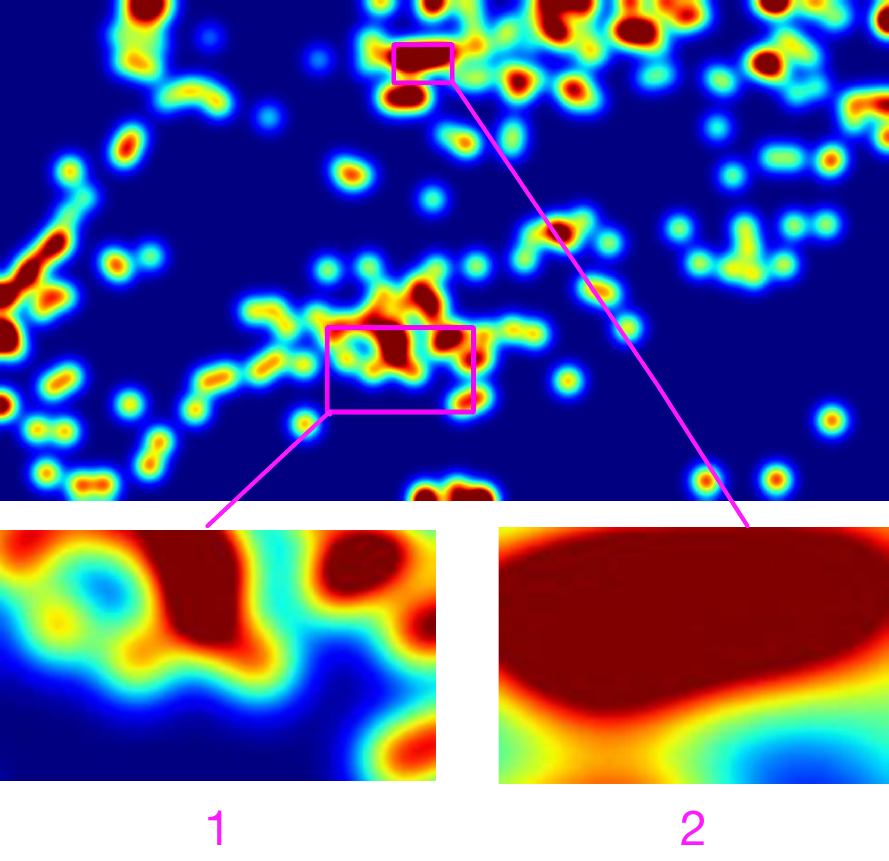}&
	\includegraphics[width=.18\linewidth]{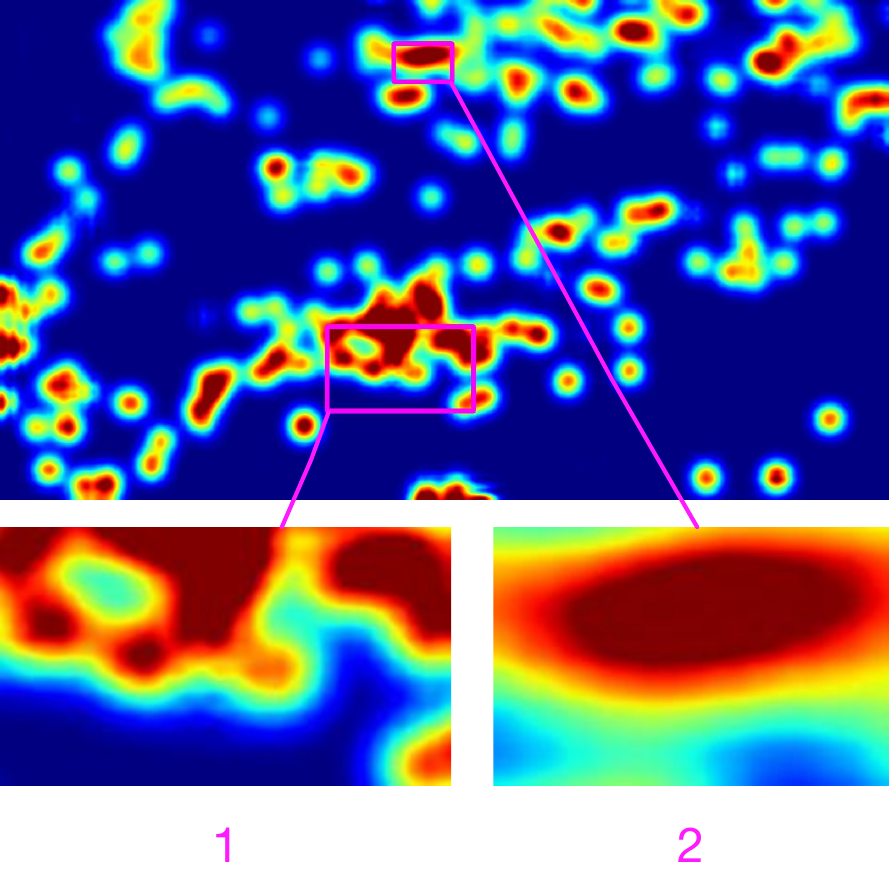}\\[-1mm]
	(a)&(b)&(c)&(d)&(e)
\end{tabular}
\vspace{-4mm}
\caption{
	{\bf Image-plane density vs ground-plane density.}  
	{\bf (a)} The two purple boxes highlight patches in which the crowd density per square meter is similar in the top image. 
	{\bf (b)} Ground-truth {\it image density} obtained by averaging the head annotations in the image plane as is done in {\it all} the approaches discussed in this paper, including ours. The bottom two patches are expanded versions of the same two purple boxes. The density appears much larger in one than in the other due to perspective distortion that increases the image density further away from the camera.
	{\bf (c)} The density estimation returned by \oursG{}.
	{\bf (d)} The ground-truth density normalized for image-scale variations so that it can be interpreted as a density per square meter. 
	{\bf (e)} The \oursG{} density similarly normalized. Note that the estimated densities in the two small windows now fall in the same range of values, which is correct. 
	}
	
\label{fig:density}
\end{figure*}

\begin{table}
  \centering
\begin{tabular}{ |p{3cm}|c|c|}
  \hline
  Model  & $MAE$ & $RMSE$  \\
  \hline
  MCNN~\cite{Zhang16s} & 145.4 & 147.3  \\
  \hline
  Switch-CNN~\cite{Sam17} & 52.8 & 59.5    \\
  \hline
  CSRNet\cite{Li18} & 35.8 & 50.0\\
  \hline
  \oursS{} & 23.5 & 38.9 \\
  \hline
  \oursG{} & \textbf{20.5} & \textbf{29.9} \\
  \hline
  \end{tabular}
  \vspace{-3mm}
  \caption{ {\bf Comparative results on the Venice dataset.} }
  \label{tab:venice}
\end{table}

\subsection{Ablation Study}


\begin{table}
  \centering
\begin{tabular}{ |p{3cm}|c|c|}
  \hline
  Model  & $MAE$ & $RMSE$  \\
  \hline
  \vggS{}  & 68.0 & 113.4 \\
  \hline
  \vggC{} & 63.4 & 108.7   \\
  \hline
  \vggN{} & 63.1 & 106.4 \\
  \hline
  \hline
  \oursS{} & \bf{62.3} & \bf{100.0}   \\
  \hline
  \end{tabular}
  \vspace{-2mm}
  \caption{{\bf Ablation study on the {\bf ShanghaiTech} part A dataset.}}
  \label{tab:ablation}
\end{table}

Finally, we perform an ablation study to confirm the benefits of encoding multiple level contextual information and of introducing contrast features. 

\parag{Concatenating and Weighting VGG Features.}
We compare our complete model without geometry, \oursS{}, against two simplified versions of it. The first one, \vggS{}, directly uses VGG-16 base features $\f_{v}$ as input to the decoder subnetwork. In other words, it does not adapt for scale. The second one, \vggC{}, concatenates all scale-aware features $\{\s_{j}\}_{1\leq j\leq S}$ to the base features instead of computing their weighted linear combination, and then passes the resulting features to the decoder. 

We compare these three methods on the {\bf ShanghaiTech} Part A, which has often been used for such ablation studies~\cite{Sindagi17,Cao18,Li18}. As can be seen in Table~\ref{tab:ablation}, concatenating the VGG features as in \vggC{} yields a significant boost, and weighing them as in  \oursS{} a further one. 

\parag{Contrast Features.}

To demonstrate the importance of using contrast features to learn the network weights, we compare \oursS{} against \vggN{} that uses the scale features ${\s_{j}}$ instead of the contrast ones to learn the weight maps. As can be seen in Table~\ref{tab:ablation}, this also results in a substantial performance loss.


\section{Conclusion and Future Perspectives}

In this paper, we have shown that encoding multi-scale context adaptively, along with providing an explicit model of perspective distortion effects as input to a deep net, substantially increases crowd counting performance. In particular, it yields much better density estimates in high-density regions.

This is of particular interest for crowd counting from mobile cameras, such as those carried by drones. In future work, we will therefore augment the image data with the information provided by the drone's inertial measurement unit to compute perspective distortions on the fly and allow monitoring from the moving drone. 

We will also expand our approach to process consecutive images simultaneously and enforce temporal consistency, which among other things implies correcting ground-truth densities to also account for perspective distortions and be able to properly reason in the terms of ground-plane densities instead of image-plane densities, which none of the approaches discussed in this paper do. We did not do it either so that our results could be properly compared to the state of the art. However, as shown in Fig.~\ref{fig:density}, the price to pay is that the estimated densities, because they are close to this image-based ground truth, need to be corrected for perspective distortion before they can be treated as ground-plane densities. An obvious improvement would therefore be to directly regress to ground densities. 

\vspace{1em}
{\bf Acknowledgments} This work was supported in part by the Swiss Federal Office for Defense Procurement.

\newpage

{\small
\bibliographystyle{ieee_fullname}
\bibliography{string,ref}

\begin{thebibliography}{10}\itemsep=-1pt

\bibitem{Arteta14}
Carlos Arteta, Victor Lempitsky, J.Alison Noble, and Andrew Zisserman.
\newblock {Interactive Object Counting}.
\newblock In {\em European Conference on Computer Vision}, 2014.

\bibitem{Arteta16}
Carlos Arteta, Victor Lempitsky, and Andrew Zisserman.
\newblock {Counting in the Wild}.
\newblock In {\em European Conference on Computer Vision}, 2016.

\bibitem{Badrinarayanan15}
Vijay Badrinarayanan, Alex Kendall, and Roberto Cipolla.
\newblock {Segnet: A Deep Convolutional Encoder-Decoder Architecture for Image
  Segmentation}.
\newblock {\em arXiv Preprint}, 2015.

\bibitem{Brostow06}
Gabriel~J. Brostow and Roberto Cipolla.
\newblock {Unsupervised Bayesian Detection of Independent Motion in Crowds}.
\newblock In {\em Conference on Computer Vision and Pattern Recognition}, pages
  594--601, 2006.

\bibitem{Cao18}
Xinkun Cao, Zhipeng Wang, Yanyun Zhao, and Fei Su.
\newblock {Scale Aggregation Network for Accurate and Efficient Crowd
  Counting}.
\newblock In {\em European Conference on Computer Vision}, 2018.

\bibitem{Chan08}
Antoni~B. Chan, Zhang-Sheng~John Liang, and Nuno Vasconcelos.
\newblock {Privacy Preserving Crowd Monitoring: Counting People Without People
  Models or Tracking}.
\newblock In {\em Conference on Computer Vision and Pattern Recognition}, 2008.

\bibitem{Chan09}
Antoni~B. Chan and Nuno Vasconcelos.
\newblock {Bayesian Poisson Regression for Crowd Counting}.
\newblock In {\em International Conference on Computer Vision}, pages 545--551,
  2009.

\bibitem{Chattopadhyay16}
Prithvijit Chattopadhyay, Ramakrishna Vedantam, Ramprasaath~R. Selvaju, Dhruv
  Batra, and Devi Parikh.
\newblock {Counting Everyday Objects in Everyday Scenes}.
\newblock In {\em Conference on Computer Vision and Pattern Recognition}, 2017.

\bibitem{Chen12f}
Ke Chen, Chen~Change Loy, Shaogang Gong, and Tao Xiang.
\newblock {Feature Mining for Localised Crowd Counting}.
\newblock In {\em British Machine Vision Conference}, page~3, 2012.

\bibitem{Fiaschi12}
Luca Fiaschi, Rahul Nair, Ullrich Koethe, and Fred~A. Hamprecht.
\newblock {Learning to Count with Regression Forest and Structured Labels}.
\newblock In {\em International Conference on Pattern Recognition}, pages
  2685--2688, 2012.

\bibitem{He14b}
Kaiming He, Xiangyu Zhang, Shaoqing Ren, and Jian Sun.
\newblock {Spatial Pyramid Pooling in Deep Convolutional Networks for Visual
  Recognition}.
\newblock In {\em European Conference on Computer Vision}, 2014.

\bibitem{He16}
Kaiming He, Xiangyu Zhang, Shaoqing Ren, and Jian Sun.
\newblock {Deep Residual Learning for Image Recognition}.
\newblock In {\em Conference on Computer Vision and Pattern Recognition}, pages
  770--778, 2016.

\bibitem{Huang17c}
Gao Huang, Zhuang Liu, Laurens van~der Maaten, and Kilian~Q. Weinberger.
\newblock {Densely Connected Convolutional Networks}.
\newblock In {\em Conference on Computer Vision and Pattern Recognition}, 2017.

\bibitem{Idrees13}
Haroon Idrees, Imran Saleemi, Cody Seibert, and Mubarak Shah.
\newblock {Multi-Source Multi-Scale Counting in Extremely Dense Crowd Images}.
\newblock In {\em Conference on Computer Vision and Pattern Recognition}, pages
  2547--2554, 2013.

\bibitem{Idrees18}
Haroon Idrees, Muhmmad Tayyab, Kishan Athrey, Dong Zhang, Somaya Al-Maadeed,
  Nasir Rajpoot, and Mubarak Shah.
\newblock {Composition Loss for Counting, Density Map Estimation and
  Localization in Dense Crowds}.
\newblock In {\em European Conference on Computer Vision}, 2018.

\bibitem{Kang18}
Di Kang and Antoni~B. Chan.
\newblock {Crowd Counting by Adaptively Fusing Predictions from an Image
  Pyramid}.
\newblock In {\em British Machine Vision Conference}, 2018.

\bibitem{Kang17}
Di Kang, Debarun Dhar, and Antoni~B. Chan.
\newblock {Incorporating Side Information by Adaptive Convolution}.
\newblock In {\em Advances in Neural Information Processing Systems}, 2017.

\bibitem{Lempitsky10}
Victor Lempitsky and Andrew Zisserman.
\newblock {Learning to Count Objects in Images}.
\newblock In {\em Advances in Neural Information Processing Systems}, 2010.

\bibitem{Li18}
Yuhong Li, Xiaofan Zhang, and Deming Chen.
\newblock {CSRNet: Dilated Convolutional Neural Networks for Understanding the
  Highly Congested Scenes}.
\newblock In {\em Conference on Computer Vision and Pattern Recognition}, 2018.

\bibitem{Lin10}
Zhe Lin and Larry~S. Davis.
\newblock {Shape-Based Human Detection and Segmentation via Hierarchical
  Part-Template Matching}.
\newblock {\em IEEE Transactions on Pattern Analysis and Machine Intelligence},
  32(4):604--618, 2010.

\bibitem{Liu18}
Jiang Liu, Chenqiang Gao, Deyu Meng, and Alexander~G. Hauptmann1.
\newblock {Decidenet: Counting Varying Density Crowds through Attention Guided
  Detection and Density Estimation}.
\newblock In {\em Conference on Computer Vision and Pattern Recognition}, 2018.

\bibitem{Liu18c}
Linbo Liu, Hongjun Wang, Guanbin Li, Wanli Ouyang, and Liang Lin.
\newblock {Crowd Counting Using Deep Recurrent Spatial-Aware Network}.
\newblock In {\em International Joint Conference on Artificial Intelligence},
  2018.

\bibitem{Liu16a}
Wei Liu, Dragomir Anguelov, Dumitru Erhan, Christian Szegedy, Scott~E. Reed,
  Cheng-Yang Fu, and Alexander~C. Berg.
\newblock {SSD: Single Shot Multibox Detector}.
\newblock In {\em European Conference on Computer Vision}, 2016.

\bibitem{Liu18b}
Xialei Liu, Joost van~de Weijer, and Andrew~D. Bagdanov.
\newblock {Leveraging Unlabeled Data for Crowd Counting by Learning to Rank}.
\newblock In {\em Conference on Computer Vision and Pattern Recognition}, 2018.

\bibitem{Long15a}
Jonathan Long, Evan Shelhamer, and Trevor Darrell.
\newblock {Fully Convolutional Networks for Semantic Segmentation}.
\newblock In {\em Conference on Computer Vision and Pattern Recognition}, 2015.

\bibitem{Onoro16}
Daniel Onoro-Rubio and Roberto~J. L{\'o}pez-Sastre.
\newblock {Towards Perspective-Free Object Counting with Deep Learning}.
\newblock In {\em European Conference on Computer Vision}, pages 615--629,
  2016.

\bibitem{Rabaud06}
Vincent Rabaud and Serge Belongie.
\newblock {Counting Crowded Moving Objects}.
\newblock In {\em Conference on Computer Vision and Pattern Recognition}, pages
  705--711, 2006.

\bibitem{Ranjan18}
Viresh Ranjan, Hieu Le, and Minh Hoai.
\newblock {Iterative Crowd Counting}.
\newblock In {\em European Conference on Computer Vision}, 2018.

\bibitem{Ren15}
Shaoqing Ren, Kaiming He, Ross Girshick, and Jian Sun.
\newblock {Faster {R-CNN}: Towards Real-Time Object Detection with Region
  Proposal Networks}.
\newblock In {\em Advances in Neural Information Processing Systems}, 2015.

\bibitem{Sam18}
Deepak~Babu Sam, Neeraj~N. Sajjan, R.~Venkatesh Babu, and Mukundhan Srinivasan.
\newblock {Divide and Grow: Capturing Huge Diversity in Crowd Images with
  Incrementally Growing CNN}.
\newblock In {\em Conference on Computer Vision and Pattern Recognition}, 2018.

\bibitem{Sam17}
Deepak~Babu Sam, Shiv Surya, and R.~Venkatesh Babu.
\newblock {Switching Convolutional Neural Network for Crowd Counting}.
\newblock In {\em Conference on Computer Vision and Pattern Recognition},
  page~6, 2017.

\bibitem{Shen18}
Zan Shen, Yi Xu, Bingbing Ni, Minsi Wang, Jianguo Hu, and Xiaokang Yang.
\newblock {Crowd Counting via Adversarial Cross-Scale Consistency Pursuit}.
\newblock In {\em Conference on Computer Vision and Pattern Recognition}, 2018.

\bibitem{Shi18}
Zenglin Shi, Le Zhang, Yun Liu, and Xiaofeng Cao.
\newblock {Crowd Counting with Deep Negative Correlation Learning}.
\newblock In {\em Conference on Computer Vision and Pattern Recognition}, 2018.

\bibitem{Simonyan15}
Karen Simonyan and Andrew Zisserman.
\newblock {Very Deep Convolutional Networks for Large-Scale Image Recognition}.
\newblock In {\em International Conference on Learning Representations}, 2015.

\bibitem{Sindagi17b}
Vishwanath~A. Sindagi and Vishal~M. Patel.
\newblock {CNN-based Cascaded Multi-task Learning of High-level Prior and
  Density Estimation for Crowd Counting}.
\newblock In {\em International Conference on Advanced Video and Signal Based
  Surveillance}, 2017.

\bibitem{Sindagi17}
Vishwanath~A. Sindagi and Vishal~M. Patel.
\newblock {Generating High-Quality Crowd Density Maps Using Contextual Pyramid
  CNNs}.
\newblock In {\em International Conference on Computer Vision}, pages
  1879--1888, 2017.

\bibitem{Szegedy15}
Christian Szegedy, Wei Liu, Yangqing Jia, Pierre Sermanet, Scott Reed, Dragomir
  Anguelov, Dumitru Erhan, Vincent Vanhoucke, and Andrew Rabinovich.
\newblock {Going Deeper with Convolutions}.
\newblock In {\em Conference on Computer Vision and Pattern Recognition}, pages
  1--9, June 2015.

\bibitem{Wang11a}
Xin Wang, Bin Wang, and Liming Zhang.
\newblock {Airport Detection in Remote Sensing Images Based on Visual
  Attention}.
\newblock In {\em International Conference on Neural Information Processing},
  2011.

\bibitem{Wu05}
Bo Wu and Ram Nevatia.
\newblock {Detection of Multiple, Partially Occluded Humans in a Single Image
  by Bayesian Combination of Edgelet Part Detectors}.
\newblock In {\em International Conference on Computer Vision}, 2005.

\bibitem{Xiong17}
Feng Xiong, Xinjian Shi, and Dit-Yan Yeung.
\newblock {Spatiotemporal Modeling for Crowd Counting in Videos}.
\newblock In {\em International Conference on Computer Vision}, pages
  5161--5169, 2017.

\bibitem{Zhang15c}
Cong Zhang, Hongsheng Li, Xiaogang Wang, and Xiaokang Yang.
\newblock {Cross-Scene Crowd Counting via Deep Convolutional Neural Networks}.
\newblock In {\em Conference on Computer Vision and Pattern Recognition}, pages
  833--841, 2015.

\bibitem{Zhang16s}
Yingying Zhang, Desen Zhou, Siqin Chen, Shenghua Gao, and Yi Ma.
\newblock {Single-Image Crowd Counting via Multi-Column Convolutional Neural
  Network}.
\newblock In {\em Conference on Computer Vision and Pattern Recognition}, pages
  589--597, 2016.

\bibitem{Zhao17b}
Hengshuang Zhao, Jianping Shi, Xiaojuan Qi, Xiaogang Wang, and Jiaya Jia.
\newblock {Pyramid Scene Parsing Network}.
\newblock In {\em Conference on Computer Vision and Pattern Recognition}, 2017.

\end{thebibliography}
}

\end{document}